\documentclass[10pt,twocolumn,letterpaper]{article}

\usepackage{cvpr}              

\usepackage[accsupp]{axessibility} 

\usepackage{graphicx}
\usepackage{booktabs}
\graphicspath{ {./figs/} }
\usepackage{subcaption}
\usepackage{xspace}
\usepackage[dvipsnames]{xcolor}
\usepackage{comment}
\usepackage{adjustbox}
\usepackage{array}
\usepackage{makecell}
\usepackage{pifont}
\usepackage{bbm}
\usepackage{multirow}
\usepackage{flushend}
\usepackage{colortbl}
\usepackage{enumitem}
\usepackage{comment}

\usepackage{sidecap}
\newcolumntype{R}[2]{%
    >{\adjustbox{angle=#1,lap=1.3\width-(#2)}\bgroup}%
    l%
    <{\egroup}%
}
\usepackage[]{pgfplots,pgfplotstable}
\usepackage[]{tikz}
\pgfplotsset{compat=1.18}

\makeatletter
\newcommand\myparagraph{\@startsection{paragraph}{4}{\z@}%
    {-6\p@ \@plus -3\p@ \@minus -3\p@}%
    {-0.5em \@plus -0.22em \@minus -0.1em}%
    {\normalfont\normalsize\itshape}}
\renewcommand\paragraph{\@startsection{paragraph}{4}{\z@}%
    {0.25ex \@plus 1ex \@minus .2ex}%
    {-1em}%
    {\normalfont \normalsize \bfseries}}
\makeatother

\definecolor{myblue}{rgb}{0.19, 0.55, 0.91}
\definecolor{myred}{rgb}{0.82, 0.1, 0.26}
\definecolor{MyGreen}{RGB}{0, 104, 55} 
\definecolor{MyRed}{RGB}{248, 3, 7} 
\definecolor{MyYellow}{RGB}{180, 180, 0}
\newcommand{\cmark}{{\textcolor{MyGreen}{\ding{51}}}}%
\newcommand{\xmark}{{\textcolor{MyRed}{\ding{55}}}}%

\def\sk{SemanticKITTI\xspace} 
\def\ns{nuScenes\xspace} 
\def\wi{WaffleIron\xspace}

\def\ours{Refiner} 

\def\ours{LOSC\xspace}
\def\BP{\Pi}
\def\prompts{T_\C}
\def\FT{\text{FT}}
\def\VLM{\text{VLM}}
\def\TBC{\text{TBC}}
\def\ABC{\text{ABC}}
\def\ATC{\text{ATC}}
\def\comb{\text{comb}}
\def\Lini{L_\text{vlm}}
\def\Labc{L_\text{abc}}
\def\Laug{L_\text{aug}}
\def\Ltim{L_\text{tim}}
\def\Lcomb{L_{\text{aug}{\otimes}\text{tim}}}
\def\Naug{N_\text{aug}}
\def\Ntim{N_\text{tim}}
\def\A{\mathcal{A}}
\def\I{\mathcal{I}}
\def\D{\mathcal{D}}
\def\C{\mathcal{C}}
\def\P{\mathcal{P}}
\def\Crobaug{\C_\text{robust-aug}}
\def\Nmin{N}
\def\taumin{\tau}
\def\ignore{\emph{ignore}}
\def\sc{$^{\smash{\text{\,(sc)}}}$}
\def\sem{$_\text{sem}$}
\def\pan{$_\text{pan}$}
\def\semsc{$_\text{sem}^\text{\,(sc)}$}
\def\pansc{$_\text{pan}^\text{\,(sc)}$}

\usepackage{pgfplots}
\usepackage[nomessages]{fp}
\usepackage{pgfplots}
\usepackage{pgfkeys}
\def\addlegendimage{\csname pgfplots@addlegendimage\endcsname}

\definecolor{cvprblue}{rgb}{0.21,0.49,0.74}
\usepackage[pagebackref,breaklinks,colorlinks,citecolor=cvprblue]{hyperref}
\usepackage[capitalize]{cleveref}

\def\paperID{86} 
\def\confName{3DV\xspace}
\def\confYear{2026\xspace}
\title{\ours: LiDAR Open-voc Segmentation Consolidator}

\author{\hspace{-3mm}Nermin Samet$^1$,
Gilles Puy$^1$,
Renaud Marlet$^{1,2}$ \\[2mm]
\hspace{-5mm}\textsuperscript{1}Valeo.ai, Paris, France
\hspace{2.5mm}\textsuperscript{2}LIGM, Ecole des Ponts, Univ Gustave Eiffel, CNRS, Marne-la-Vall\'ee, France \\
}

\begin{document}
\maketitle

\begin{abstract}
We study the use of image-based Vision-Language Models (VLMs) for open-vocabulary segmentation of lidar scans in driving settings. Classically, image semantics can be back-projected onto 3D point clouds. Yet, resulting point labels are noisy and sparse. We consolidate these labels to enforce both spatio-temporal consistency and robustness to image-level augmentations. We then train a 3D network based on these refined labels. This simple method, called LOSC, outperforms the SOTA of zero-shot open-vocabulary semantic and panoptic segmentation on both nuScenes and SemanticKITTI, with significant margins. Code is available at \url{https://github.com/valeoai/LOSC}.
\end{abstract}

\section{Introduction}

Vision-Language Models (VLMs)  are flourishing, offering unprecedented capabilities \cite{zhang2024vlmsurvey, wu2024openvocsurvey, zhou2024imagesegmentationfoundationmodel, li2025surveystateartlarge}: employed as general Vision Foundation Models (VFMs), they are usable off-the-shelf for many open-vocabulary tasks: image classification, visual reasoning, visual question answering, image retrieval, image captioning, object detection, semantic segmentation, panoptic segmentation, etc. However, for a number of closed-set and domain-specific tasks, including object detection and semantic or instance segmentation, 
zero-shot performance of VLMs is not up to the level of supervised finetuning with manual labels~\cite{feng2025vlmobjectdetection}. Several reasons may explain it.

First, some VLMs are trained with the sole supervision of image-text pairs, that makes dense tasks as segmentation harder to learn \cite{radford2021clip}. Second, as training a VLM requires lots of data, some models are trained on data scrapped from the Web, which are thus subject to a distribution bias: they contain comparatively few samples from particular domains, e.g., medical images \cite{schuhmann2002laion5b, chen2025reproducible}. It concerns both the visual contents (images) and the associated vocabulary (text). Third, ``vision'' in ``VLM'' generally only implies images, as 3D data (depth maps, point clouds, etc\mbox{.}) aligned with text is in much shorter supply than 2D data \cite{ma2024llmsstep3dworld}.

\begin{figure}
\centering
\begin{minipage}{0.885\linewidth}
\centering
\resizebox{0.885\linewidth}{!}{
\begin{tikzpicture}[baseline={([yshift=\baselineskip]current bounding box.north)}]
    \tikzstyle{every node}=[font=\small]
    \begin{axis}[
        ymin=0,
        width=5cm,
        height=6cm,
        x=0.9cm, 
        enlarge x limits={abs=0.5cm}, 
        xtick={1,2,3,4,6,7},
        x tick style={draw=none},
        x tick label style={font=\footnotesize},
        xticklabels={},
        y tick label style={font=\footnotesize},
        ymajorgrids = true,
        every axis plot/.append style={
          ybar,
          bar width=.6cm,
          bar shift=0pt,
          fill,
        },
        title={\hspace{8mm} \ns \hspace{23mm} \sk},
        title style={yshift=-1.ex,},
        legend pos=south east,
        grid=major,
        legend style={nodes={scale=0.8, transform shape}},
    ]
    \addplot[color=Purple!40, ybar legend] coordinates {
        (1, 45.5)  %
    };
    \addplot[color=Purple!60] coordinates {
        (2, 46.1) %
    };    
    \addplot[color=Purple] coordinates {
        (3, 47.9) 
    };
    \addplot[color=ForestGreen!60] coordinates {
        (4, 49.3) 
    };
    \addplot[color=Purple] coordinates {
        (6, 28.7) 
    };
    \addplot[color=ForestGreen!60] coordinates {
        (7, 35.2) 
    };
    \addplot[style={
        solid,
        thick,
        mark=none,
        mark options={solid},
        smooth, 
        Purple!60!black,
        dashed,
    }]
    coordinates {(0.5, 49.3) (4.5, 49.3)};
    \addplot[style={
        solid,
        thick,
        mark=none,
        mark options={solid},
        smooth, 
        Purple!60!black,
        dashed,
    }]
    coordinates {(5.5, 35.2) (7.5, 35.2)};    
    \coordinate (LClipScene) at (0.45cm,2.0cm);
    \coordinate (dClipScene) at (0.4cm,4.2cm);
    \coordinate (LSAL) at (1.35cm,2.0cm);
    \coordinate (dLSAL) at (1.3cm,4.2cm);
    \coordinate (LOpenScene) at (2.25cm,2.0cm);
    \coordinate (dOpenScene) at (2.2cm,4.2cm);
    \coordinate (LLOSC) at (3.15cm,2.0cm);    
    \coordinate (dLOSC) at (3.1cm,4.2cm);
    \coordinate (KITTILSAL) at (4.95cm,1.2cm);
    \coordinate (KITTIdSAL) at (4.9cm,3.1cm);
    \coordinate (KITTILLOSC) at (5.85cm,1.5cm);    
    \coordinate (KITTIdLOSC) at (5.8cm,3.1cm);
    \end{axis}
    \node[rotate=90] at (LClipScene) {\textcolor{white}{\texttt{\footnotesize OV3D~[19]}}};
    \node[rotate=90] at (LSAL) {\textcolor{white}{\texttt{\footnotesize GGSD [16]}}};
    \node[rotate=90] at (LOpenScene) {\textcolor{white}{\texttt{\footnotesize AFOV [17]}}};
    \node[rotate=90] at (LLOSC) {\textcolor{white}{\texttt{\footnotesize \ours (ours)}}};
    \node[] at (dClipScene) {\textcolor{Purple!60!black}{\scriptsize{-3.8}}};
    \node[] at (dLSAL) {\textcolor{Purple!60!black}{\scriptsize{-3.2}}};
    \node[] at (dOpenScene) {\textcolor{Purple}{\scriptsize{-1.4}}};
    \node[] at (dLOSC) {\textcolor{ForestGreen!60!black}{\scriptsize{49.3}}};
    \node[rotate=90] at (KITTILSAL) {\textcolor{white}{\texttt{\footnotesize SAL [22]}}};
    \node[rotate=90] at (KITTILLOSC) {\textcolor{white}{\texttt{\footnotesize \ours (ours)}}};
    \node[] at (KITTIdLOSC) {\textcolor{ForestGreen!60!black}{\scriptsize{35.2}}};
    \node[] at (KITTIdSAL) {\textcolor{Purple}{\scriptsize{-6.5}}};
\end{tikzpicture}
}
\\
\hspace{8mm}Semantic segmentation (mIoU\%)
\end{minipage}
\begin{minipage}{0.885\linewidth}
\centering
\resizebox{0.885\linewidth}{!}{
\begin{tikzpicture}[baseline={([yshift=\baselineskip, xshift=2pt]current bounding box.north)}]
    \tikzstyle{every node}=[font=\small]
    \begin{axis}[
        ymin=0,
        width=5cm,
        height=6cm,
        x=0.9cm, 
        enlarge x limits={abs=0.5cm}, 
        xtick={1,2,4,5,6},
        x tick style={draw=none},
        x tick label style={font=\footnotesize},
        xticklabels={},
        y tick label style={font=\footnotesize},
        ymajorgrids = true,
        every axis plot/.append style={
          ybar,
          bar width=.6cm,
          bar shift=0pt,
          fill
        },
        title={\ns \hspace{14mm} \sk},
        title style={yshift=-1.ex,},
        legend pos=south east,
        grid=major,
        legend style={nodes={scale=0.8, transform shape}},
        legend cell align={left},
    ]
    \addplot[color=MidnightBlue!80] coordinates {
        (1, 38.4) 
    };
    \addplot[color=Peach!80] coordinates {
        (2, 48.4) 
    };
    \addplot[color=MidnightBlue!60] coordinates {
        (4, 24.8) 
    };
    \addplot[color=MidnightBlue!80] coordinates {
        (5, 30.8) 
    };
    \addplot[color=Peach!80] coordinates {
        (6, 32.4) 
    };
    \addplot[style={
        solid,
        thick,
        mark=none,
        mark options={solid},
        smooth, 
        MidnightBlue!80!black,
        dashed,
    }]
    coordinates {(0.5, 48.4) (2.5, 48.4)};
    \addplot[style={
        solid,
        thick,
        mark=none,
        mark options={solid},
        smooth, 
        MidnightBlue!80!black,
        dashed,
    }]
    coordinates {(3.5, 32.4) (6.5, 32.4)};
    \coordinate (LSAL) at (0.45cm,1.7cm);
    \coordinate (dSAL) at (0.4cm,4.2cm);
    \coordinate (LLOSC) at (1.35cm,2.1cm);    
    \coordinate (dLOSC) at (1.3cm,4.2cm);
    \coordinate (KITTILSAL) at (3.15cm,1.0cm);
    \coordinate (KITTIdSAL) at (3.1cm,2.9cm);
    \coordinate (KITTILSAL4D) at (4.05cm,1.3cm);
    \coordinate (KITTIdSAL4D) at (4.0cm,2.9cm);
    \coordinate (KITTILLOSC) at (4.95cm,1.4cm);    
    \coordinate (KITTIdLOSC) at (4.9cm,2.9cm);
    \end{axis}
    \node[rotate=90] at (LSAL) {\textcolor{white}{\texttt{\footnotesize SAL [22]}}};
    \node[rotate=90] at (LLOSC) {\textcolor{white}{\texttt{\footnotesize \ours (ours)}}};
    \node[] at (dSAL) {\textcolor{MidnightBlue!80!black}{\scriptsize{-10.0}}};
    \node[] at (dLOSC) {\textcolor{Peach!80!black}{\scriptsize{48.4}}};
    \node[rotate=90] at (KITTILSAL) {\textcolor{white}{\texttt{\footnotesize SAL [22]}}};
    \node[rotate=90] at (KITTILSAL4D) {\textcolor{white}{\texttt{\footnotesize SAL-4D [24]}}};
    \node[rotate=90] at (KITTILLOSC) {\textcolor{white}{\texttt{\footnotesize \ours (ours)}}};
    \node[] at (KITTIdSAL) {\textcolor{MidnightBlue!80!black}{\scriptsize{-7.6}}};
    \node[] at (KITTIdSAL4D) {\textcolor{MidnightBlue!80!black}{\scriptsize{-1.6}}};
    \node[] at (KITTIdLOSC) {\textcolor{Peach!80!black}{\scriptsize{32.4}}};
\end{tikzpicture}
}
\\
\hspace{8mm}Panoptic segmentation (PQ\%)
\end{minipage}
\caption{\textbf{Performance of \ours compared to SOTA on zero-shot open-voc segmentation.}}
\label{fig:teaser}
\end{figure}
Therefore, most existing 3D VLMs (relating 3D data and text) actually operate via intermediate 2D data, leveraging a 2D VLM already trained on a large quantity of images: the 2D VLM is used to extract 2D information (labels or features), which is then back-projected onto 3D~\cite{ma2024llmsstep3dworld}. The 3D labels (or features) can then be directly employed as such, or used as pseudo-labels to train a 3D model, e.g., for the feature-based distillation of a 2D VFM into a 3D VFM \cite{scalr}.

In this paper, we study the use of 2D VLMs for the zero-shot open-vocabulary segmentation of lidar scans in driving settings, transferring 2D knowledge to 3D while specializing it for a specific purpose. Precisely, given (i)~\emph{an open-vocabulary 2D VLM}, (ii)~\emph{any set of classes}, defined via \emph{arbitrary textual prompts}, and (iii)~\emph{an unannotated 3D dataset} of lidar scans with registered images, we create \emph{a 3D model} that efficiently can infer the semantic segmentation of any lidar scan for these specific classes. Although this 3D model is \emph{closed-set}, its specification, based on textual prompts, is \emph{fully open-set}, and the process of creating the 3D model is \emph{annotation-free}. Besides, as we can leverage a state-of-the-art (SOTA) 3D segmenter, which we train from the segments produced by the 2D VLM, we obtain an efficient 3D model.

However, driving scenes suffers from the data shortage mentioned above: for this specific domain, there is relatively little visual content available associated with text \cite{schuhmann2002laion5b, chen2025reproducible}, and even less involving 3D data \cite{ma2024llmsstep3dworld}. As a result, current VLMs still lag behind full supervision for the semantic and panoptic segmentation of driving scene images, and even more so for lidar scans
\cite{feng2025vlmobjectdetection, akter2025imagesegmentationllmsurvey, wu2024ptv3}.
Concretely, in driving scenes, VLM-based image segments  are often missing or wrongly labeled, while object boundaries are imprecise. This label inaccuracy transfers to 3D point after back-projection.
Additionally, some 3D points may get wrong 2D information due to parallax between camera and lidar sensors, and 3D points invisible from cameras (not in their field of view) do not get any information.

We propose three label consolidation mechanisms to address these issues. First, we consider that semantics should be robust to basic image augmentations and discard labels that change according such variations. Second, we assume that 3D labeling should be consistent over time while driving, at least regarding the static part of the scene. Therefore, we ensure that all points falling into the same voxel get the same label over a whole driving sequence. Third, it is known that iterated training has a denoising effect on pseudo-labels. Therefore, starting from a (self-supervised) pretrained 3D model, we run several finetuning iterations, interleaving them with spatio-temporal consolidation.

To get the best performance, we additionally (i)~benchmark existing VLMs for semantic segmentation, (ii)~experiment with different prompt variants, and (iii)~leverage a pretrained 3D backbone. It allows us to improve the state of the art (SOTA) of open-vocabulary 3D semantic segmentation by a significant margin on two classic datasets
(see \cref{fig:teaser}).
Our contributions are as follows:
\begin{itemize}
    \item 
    We propose a way to combine several techniques to improve VML-based 3D segmentation, including augmentation- and time-based consolidations, 3D model pretraining and iterated finetuning.
    \item 
    We additionally study the specific choice of a basis VLM for the driving setting, for both semantic and panoptic segmentation, using different metrics and prompts.
    \item 
    Despite the simplicity of our method 
    (no complex training or architecture, no prompt engineering or VLM tweaking, very few and little-sensitive parameters), we largely improve the SOTA of open-vocabulary semantic and panoptic segmentation of lidar data.
\end{itemize}

\section{Related work}
\label{sec:rw}

\paragraph{Open-vocabulary 2D segmentation.} 
Models such as CLIP \cite{radford2021clip} and ALIGN \cite{jia2021align}, which aligns global text and image representations, demonstrate strong capabilities in zero-shot classification. 
These capabilities have been extended to dense prediction tasks requiring pixel-level information, such as semantic segmentation. 
Some methods leverage the original CLIP model and find recipes to extract dense pixel representations from it. 
For example, MaskCLIP~\cite{zhou2022maskclip} extract patch features in the last attention layer. 
CLIP-DIY~\cite{clipdiy} aggregates features from several crops.
Some other methods finetune or train from scratch a CLIP-like architecture but where text representations are aligned with pixel- or patch-level representations \cite{li2022lseg,openseg,seem,openseed,sam,regionclip}. In this work, we evaluate the following models: OpenSeg~\cite{openseg}, SEEM~\cite{seem}, OpenSeed~\cite{openseed} and Grounded-SAM~\cite{sam}. 

\paragraph{Open-vocabulary 3D segmentation.}
The capabilities of open-vocabulary 2D semantic segmentation can be transferred to 3D network by distillation.

OpenScene~\cite{openscene} and CLIP2Scene~\cite{clip2scene} do so using point-pixel correspondences and training a 3D network to produce point features which align with the corresponding pixel representations extracted from a CLIP-like model. The process was then improved by exploiting, e.g., additional geometric regularization~\cite{wang2024ggsd,sunaaai25}, improving text queries~\cite{autovoc,jiang2024ov3d}, or exploiting models such as SAM~\cite{sam} for denoising~\cite{chen2023towards,zou2025adaco}.

SAL~\cite{sal} uses SAM~\cite{sam} to get instance masks from images, and MaskCLIP~\cite{ding2023maskclip} to get per-mask CLIP features. The masks and features are lifted in 3D with point-pixel correspondences, and a 3D network is trained to predict the pairs of masks and features. SAL therefore enables zero-shot lidar panoptic segmentation. SAL was recently improved by operating on sequences rather than single scans in \cite{sal4d}. In this work, we train our 3D network for semantic segmentation and leverage ALPINE~\cite{alpine} to obtain panoptic labels.

LeAP~\cite{leap}, developed concurrently, is closer to our method. LeAP leverages a VLM to generate 2D soft labels, which are lifted in 3D. The 3D soft labels are refined using a Bayesian update leveraging time consistency. To further improve the quality of these 3D soft labels, a 3D network is trained on the most confident soft labels and the output of this trained 3D network is used to enrich the 3D soft labels originating for images. 
Unlike LeAP, we exploit one-hot labels instead of soft-labels and a simple majority voting when leveraging time-consistency, instead of a Bayesian update. But more importantly, we provide new insights by (a)~comparing the performance of several VLMs, (b)~showing how to best combine image augmentations and time-consistency to improve pseudo-label quality, and (c)~demonstrating that using a pretrained 3D network gives a significant performance boost.

\section{Method}
\label{sec:method}
\begin{figure*}
\centering
\includegraphics[width=1.\textwidth]{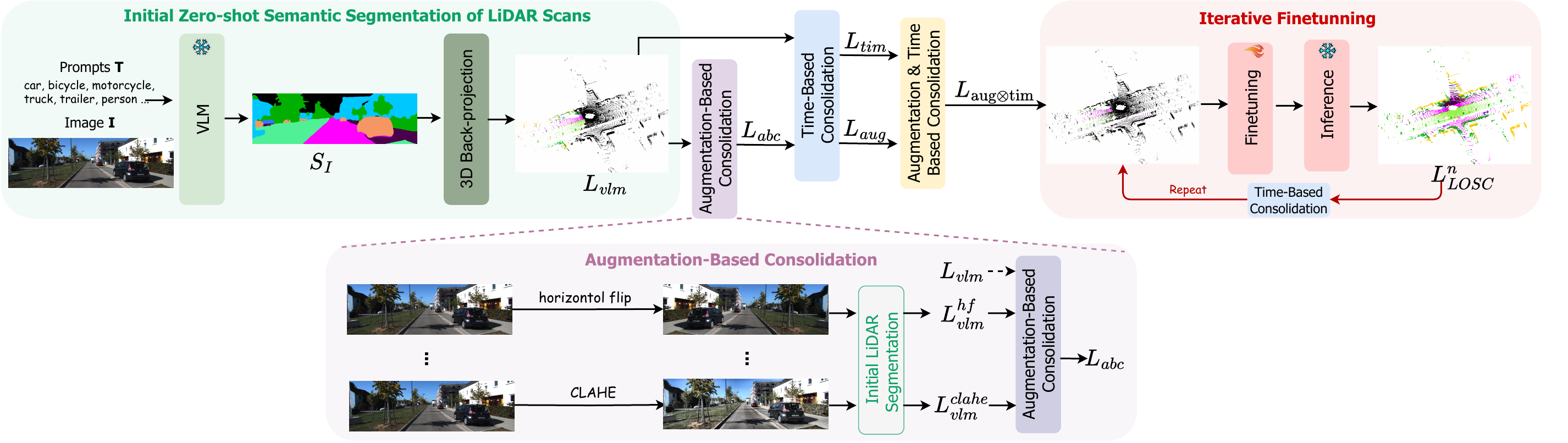}
\caption{\textbf{Overall pipeline of \ours.} The images are first segmented using an open-vocabulary segmentation model. The semantic 2D labels are then backprojected on the lidar points. These 3D labels are refined using three consolidation steps: the first uses label consistency across several image augmentation, the second uses time consistency, the last combines the best labels from the previous steps. These labels are used to finetune a 3D network. Finally, few steps of self-training are used to improve the results.}
\label{fig:base_model}
\end{figure*}

Our goal is the zero-shot open-voc labeling of lidar scans. We assume we are given (i)~a closed but unconstrained set of classes $\C$, (ii)~a VLM providing semantic segments of images from open-vocabulary textual prompts, (iii)~an unlabeled training dataset $\D$ of sequences of synchronized and calibrated images and scans.
The overall pipeline of our method, illustrated on \cref{fig:base_model}, is as follows: 
\begin{enumerate}
\item
Based on prompts for the target set of classes, we use the VLM to produce a semantic segmentation of the dataset of images. Back-projecting these 2D segments onto the associated point clouds then provides initial, basic semantic labels for 3D points. 
\item
These initial 3D semantic labels are first consolidated using a temporal aggregation of lidar scans in a sequence, and voxel-level majority voting. 
\item
Considering a set of augmentations applied to images before they go through the VLM, we consolidate 3D point labels by only keeping those that remain consistent.  
\item
Next, we combine augmentation-based consolidation and time-based consolidation, ensuring that label classes are represented enough to be used for training. 
\item
Finally, the resulting labels are used to finetune a pretrained 3D network. Iterating a few times temporal voxel consolidation and finetuning further improves segmentation quality. 
\end{enumerate}
Each of these steps is detailed in the following subsections.

\subsection{Initial Zero-shot Semantic Segmentation of LiDAR data with the VLM}
\label{sec:raw_labels}

Formally, given an image $I$ and textual prompts $\prompts$ for the target classes $\C$, the VLM produces a 2D pixelwise semantic segmentation $S_I = \VLM(I,\prompts)$. It is back-projected into the 3D points of the associated lidar point cloud $P$ to provide a 3D pointwise semantic segmentation $S_P = \BP_{I,P}(S_I)$, where $\BP_{I,P}$ denotes the back-projection. 3D points falling outside of the camera frustrum, or hidden behind other points due to parallax configurations between camera and lidar, get the $\ignore$ label. When a lidar scan $P$ is captured together with several images $\I(P)$, the images are segmented and back-projected jointly, forming the initial VLM-based point cloud labeling $\Lini$\rlap:
\begin{align}
    \Lini^P &= \cup_{I \in \I(P)}\; \BP_{I,P}(\VLM(I,\prompts))
\end{align}
(In the following, we drop subscripts, superscripts and parameters if they are clear from the context.)
These initial labels are both noisy (incorrect VLM labeling, camera-lidar parallax inconsistencies) and sparse (missing 2D VLM labeling, 3D points invisible from cameras) in the sense that a point $p \in P$ is given a label $\Lini(p) \in \C \cup \{\ignore\}$, where $\ignore$ represents the lack of semantics. For a dataset like SemanticKITTI \cite{semantickitti}, this sparsity is particularly severe (cf.\,\cref{sec:experiments}), because the 360$^\circ$ lidar is associated to only one camera, with a 90$^\circ$ field of view. In the following, we introduce three refinement strategies to improve both label quality and quantity.

\subsection{Label Refinement with Time-Based Consolidation (\TBC)}
\label{sec:tempovox}

Due to resolution, faraway objects are hard to identify as they are perceived with only few pixels and points. However, as the ego vehicle drives and comes closer to them, their semantics becomes more apparent. To leverage semantic consistency over time, we aggregate each sequence $\P$ of labeled point clouds into the same coordinate system and enforce label consistency at voxel level by majority voting. Then, for any individual scan $P \in \P$, we assign to each point $p \in P$ the label of the voxel $p$ falls into. It produces scans $\Ltim^\P = \TBC(\Lini^\P)$ with time-consistent labels.

This process may increase label sparsity because, in our implementation, $\ignore$ is treated as a full-fledged class label. The idea is that a majority of points in a voxel vote for $\ignore$, then all points falling into that voxel should be considered as unreliable and labeled as $\ignore$ too. Besides, even when such a configuration does not arise, the label distribution may also change, with some classes loosing point representatives in favor of other classes.

A variant consists in weighting the label votes with the softmax values of the label prediction. However, empirically, the gain is only marginal and it is simpler just to count class labels in voxels.

\subsection{Label Refinement with Augmentation-Based Consolidation (\ABC)}
\label{sec:augmentations}

A robust network should be little sensitive to small input variations. To filter out ``brittle'' labels, we probe VLM labeling confidence using image augmentations: a label is considered as untrustful if it changes across a set of image augmentations. This Augmentation-Based Consolidation (\ABC) can be seen as a form of Test-Time Augmentation (TTA) applied to VLM inference.

Formally, we consider a set of image augmentations $\A$. For any augmentation $A \in \A$, we denote by $\Lini^A$ the initial point cloud labels produced by the VLM from augmented images $A(\I(P))$. If the label of a point $p \in P$ is the same class $c$ whatever the augmentation $A \in \A$, then the augmentation-based consolidated label $\Laug(p)$ of $p$ is this class $c$; otherwise, it is the $\ignore$ label:
\begin{align}
\Lini^{A} &= \bigcup_{I \in \I} \; \BP_{I}(\VLM(A(I), \prompts)) \\
\Labc(p) &= \ABC(\Lini^\A)(p) \nonumber \\
         &= \begin{cases}
c, & \text{if } \forall A \in \A,\; \Lini^A(p) = c \\
\ignore, & \text{otherwise}
\end{cases}
\label{eq:Labc}
\end{align}

We then enforce temporal voxel consistency and build a consolidated labeling $\Laug = \TBC(\Labc)$.

\subsection{Label Refinement Combining Augmentation- and Time-based Consolidation (\ATC)}
\label{sec:combine}

Even more than time-based consolidation, augmentation-based consolidation tends to increase label quality but also to reduce label quantity (cf.\ \cref{sec:experiments}). To make sure points of each class remain both numerous enough and balanced enough, we introduce a way to combine $\Laug$ and $\Ltim$.

Let $N^c = |\{L(p)=c\}_{p\in P}|$ be the number of points in a labeling $L$ of point cloud $P$ that have been given label~$c$. The set of classes in $\C$ that are robust to augmentation-based consolidation is
\begin{align}
    \Crobaug &= \{c \in \C \mid \Naug^c \geq \Nmin \text{~and~} \Naug^c / \Ntim^c \geq \taumin \}
\end{align}
for some minimum number of points $\Nmin$ and minimum proportion of preserved points $\taumin$.
For these classes, it is safe to use the augmentation-based labeling $\Laug$. Otherwise, it is safer to stick to time-based labeling $\Ltim$. Therefore, we construct the following combined labeling $\Lcomb$:

\begin{align}
\Lcomb(p) &= \comb(\Laug,\Ltim)(p) \nonumber \\
          &= \begin{cases}
              \Laug(p), & \Laug(p) \in \Crobaug \\
              \Ltim(p), & \text{otherwise}
             \end{cases}
\end{align}

This hybrid strategy helps preserve class balance and label quality, especially for rare classes and for categories that are hard to predict.

\subsection{Label Refinement with Iterative FineTuning (FT)}
\label{sec:iterativeft}

Finally, we use the combined consolidated labels of $\Lcomb$, obtained on the whole training set $\D$, as pseudo-labels to train a 3D network. While all previous labelings ($\Lini$, $\Ltim$, $\Laug$, $\Lcomb$) are partial, in the sense that some points may be labeled as $\ignore$, applying the trained 3D network on any scan yields a fully-labeled point cloud (no $\ignore$).

To improve the performance of this 3D segmenter, we do two things. First, we actually start from a pretrained model $M_0$ and finetune it (\FT) on the pseudo-labels of $\Lcomb(\D)$, resulting in a model $M_1 = \FT_{\Lcomb(\D)}(M_0)$. Second, we iterate training, as often done in semi-supervised settings. More precisely, given a trained model $M_n$, we create new pseudo-labels $M_n(\D)$ for the training set, apply time-based consolidation, and use these refined pseudo-labels to finetune $M_n$, resulting a new model $M_{n+1} = \FT_{\TBC(M_n(\D))}(M_n)$.
Please note that augmentation-based consolidation (\ABC) is performed only once and used via $\Lcomb$ to finetune $M_1$. For all subsequent iterations, only time-based consolidation is used, to refine pseudo-labels of the training set before the next finetuning.
The resulting model, after a few iterations, is our final model, which we refer to as \ours.

\section{Experiments}
\label{sec:experiments}

\begin{table}[!t]
\centering
\setlength\extrarowheight{-0.5pt}
\resizebox{0.49\textwidth}{!}{
\begin{tabular}{lccccccc}
\toprule

\multirow{3}{*}{Method}  &  \multicolumn{2}{c}{Segmentation} &  \multicolumn{2}{c}{Prompt} & \multirow{3}{*}{\bf \shortstack{Label \\ cov.\% }}  & \multirow{2}{*}{\bf mAcc}   & \multirow{2}{*}{\bf mIoU}   \\
 \cmidrule(lr){2-3}  \cmidrule(lr){4-5}
& Sem.  &   Pan. &  Min.  & Rich &   &  \textbf{\%}  & \textbf{\% } \\
\midrule
GSAM~\cite{gsam} & \cmark & \xmark &  \cmark & \xmark & 68.0 &  32.8 & 21.1 \\ 
\midrule
GSAM 2~\cite{gsam}  & \cmark & \xmark &  \cmark & \xmark & 75.3 &  46.6 & 24.3    \\ 
\midrule
\multirow{4}{*}{OpenSeeD~\cite{openseed}} & \cmark & \xmark & \cmark & \xmark & 74.9 & 43.4 & 33.4 \\ 
 & \cmark & \xmark &  \xmark &  \cmark & 74.8 & 42.8 & 33.8  \\ 
 & \xmark & \cmark &  \cmark & \xmark & 63.9 & 39.5 &31.9  \\ 
 & \xmark & \cmark &  \xmark &  \cmark & 66.9 &  39.9 &32.4 \\ 

\midrule
\multirow{4}{*}{SEEM~\cite{seem}} & \cmark & \xmark &  \cmark &  \xmark & 76.2 & 47.7 & 29.6 \\ 
 & \cmark & \xmark &  \xmark &  \cmark & 76.3 & 44.7 & 25.7   \\ 
 & \xmark & \cmark &  \cmark & \xmark & 39.9 & 25.2 & 21.1 \\ 
 & \xmark & \cmark &  \xmark &  \cmark & 23.3 &  13.1 & 8.3 \\ 

\midrule
\multirow{2}{*}{OpenSeg~\cite{openseg}} & \cmark & \xmark &  \cmark &  \xmark & 72.8 & 46.5& 26.7\\ 
 & \cmark & \xmark &  \xmark &  \cmark & 76.3 & 43.7 &  24.0 \\ 

\bottomrule
\end{tabular}
}
\caption{\textbf{Semantic segmentation benchmark of 2D open-vocabulary VLMs} on the \ns validation set. Performance is evaluated using minimal or rich prompts, with semantic or (merged) panoptic segmentation results where applicable. \textit{Label cov.} is label coverage.}
\label{table:2d_open_voc}
    
\end{table}

In this section, we analyze our proposed method and compare it against the current SOTA.

\paragraph{Implementation details.}

On the image side, for augmentation-based consolidation, we use the Albumentations library~\cite{albumentations}, applying 10 distinct image transformations: horizontal-flip, hue-saturation, blur, color-jitter, auto-contrast, sharpen, chromatic-aberration, emboss, fancy-pca, clahe. The appendix details the rationale of this choice.
This set of augmentations is consistent in the sense that although a single disagreement is enough to discard the label of a point (cf.\,\cref{eq:Labc}), empirical results show the robustness of pseudo-labeling (cf.\ appendix), leading to SOTA results (cf.\,\cref{table:sota_semantic_new}).

Regarding 3D model pretraining, we leverage a \emph{self-supervised} backbone to benefit from class-agnostic representations for our open-set segmentation approach.
As 3D network, we thus chose WaffleIron~\cite{waffleiron} (as did LeAP \cite{leap}), more precisely WI-48-768, for which self-supervised pretrained weights with ScaLR~\cite{scalr} are publicly available. This pretrained model is currently among SOTA self-supervised 3D backbones for lidars.
When finetuning this network, we use the protocol of~\cite{scalr} but reduce the number of epochs to 10. 
For trainings without ScaLR initialization, we train the model from scratch as described in~\cite{waffleiron}. 

For spatio-temporal consolidation, we set the voxel size to 10~cm. 
For the combination of  consolidations, we set $\Nmin = 200k$ and $\taumin = 1/3$.
Last, we limit ourselves to 3 finetuning iterations, which defines our method, \ours.
We experiment on both nuScenes~\cite{nuscenes} and SemanticKITTI~\cite{semantickitti} with identical parameters. 

\paragraph{Benchmarking 2D VLMs.}
We evaluate the quality of 3D semantic labels $\Lini$ obtained from several established 2D open-vocabulary VLMs after back-projection: Grounded SAM~\cite{gsam}, SEEM~\cite{seem}, OpenSeeD~\cite{openseed} and OpenSeg~\cite{openseg}.
We use two types of prompts in our experiments: minimal and rich prompts (which are detailed in the Appendix).
Some VLMs can output either semantic or panoptic masks. We tested both alternatives: panoptic masks were turned into semantic masks by simply merging all instance-masks from each class.
The quality of $\Lini$ labels is evaluated on the validation set on nuScenes. We compute the label coverage (percentage of annotated points after back-projection of annotated pixels) as well as the mAcc and mIoU, where non-annotated points after back-projection are counted as errors.
The results are presented in Table~\ref{table:2d_open_voc}.
\begin{table*}[!t]
\centering
\resizebox{0.8\textwidth}{!}{
\begin{tabular}{l c c  c c  c c}
\toprule
& Time & Augmentation & ScaLR & \ns & \sk \\
Labeling  &  consistency  & consistency & pretraining   & mIoU\% & mIoU\% \\
\midrule
\multirow{2}{*}{$\Lini = \BP(\VLM(\I,\prompts))$} & \xmark & \xmark &  \xmark & 37.4 & \hphantom{0}6.5 \\ 
& \xmark & \xmark &  \cmark & 40.5 & 22.6 \\ 
\midrule
$\Ltim = \TBC(\Lini)$ & \cmark & \xmark &  \cmark & 46.4 & 32.1 \\ 
$\Labc = \ABC(\Lini)$ & \xmark & \cmark &  \cmark & 39.2 & 23.0 \\ 
$\Laug = \TBC(\ABC(\Lini))$ & \cmark & \cmark &  \cmark & 46.5 & 31.1 \\ 
\midrule
\multirow{2}{*}{$\Lcomb = \comb(\Laug,\Ltim)$} & \cmark & \cmark &  \xmark & 45.7 & 31.0 \\ 
 & \cmark & \cmark &  \cmark & 48.0 & 34.2 \\ 
\bottomrule
\end{tabular}
}
\caption{\textbf{Effect of our label consolidations} on \ns and \sk validation sets.}
\label{tab:nssk-ablation}
    
\end{table*}

First, we notice that about 65 to 75\% of the points receive a label in most configurations, with the exception of the panoptic variant of SEEM where less than 40\% of the points get a label. The differences in label coverage is explained by the fact that the VLMs do not provide labels for some pixels.
Second, our results show moderate 
performance differences between the minimal and rich prompts, especially for OpenSeeD, but except for SEEM. 
For the Grounded SAM family of models, we report results only with minimal prompts as the runtime increases substantially with more elaborate prompts. 
Third, we notice that the semantic variant of VLMs tends to provide labels of better quality than the (merged) panoptic variant.
Finally, we notice that OpenSeeD in the best performing model regarding mIoU.
Based on these results, we generate $\Lini$ for the \emph{training sets} of \sk and \ns by back-projecting 2D labels obtained using the semantic variant of OpenSeeD with minimal prompts, i.e., the fastest, simplest and nearly best discovered configuration.

\paragraph{3D model pretraining and label refinement.}

Table~\ref{tab:nssk-ablation} shows
the benefits 
of leveraging a pretrained 3D network and of using our label refinement strategies.

First, we observe that training from ScaLR pretrained weights, compared to random weights, significantly boosts performance by 3.1 and 16.1 mIoU points on \ns and \sk, respectively, when starting from $\Lini$.  
Starting from pretrained weights continues to improve performance even when starting from our most refined pseudo-labels $\Lcomb$.

Second, applying our time-based consolidation strategy ($\TBC$) directly on $\Lini$, producing $\Ltim$, further improves performance by about 6 and 10 points on \ns and \sk, respectively.

Third, our augmentation-based consolidation ($\ABC$), directly applied on $\Lini$ to produce $\Labc$, brings little gain on \sk over just training from $\Lini$, and even slightly degrades performance on \ns. Besides, refining $\Labc$
with $\TBC$ to produce $\Laug$ leads to similar results than with just $\Ltim$. 
But ABC is actually not meant to operate on it own. The benefit of augmentation-based consolidation appears in fact when combining $\Laug$ and $\Ltim$, where we obtain an extra boost of performance of 1.6 and 2.1  
mIoU points on \ns and \sk, respectively, compared to using $\Ltim$ alone. 
\begin{table*}[!t]
\centering
\resizebox{0.8\textwidth}{!}{
\begin{tabular}{l c c  c | c c  c  }
\toprule

&  \multicolumn{3}{c}{\ns~~(mIoU\%)} &  \multicolumn{3}{c}{\sk~~(mIoU\%)} \\
 \cmidrule(lr){2-4}  \cmidrule(lr){5-7}
 Labeling & 
 1\textsuperscript{st} iter. & 2\textsuperscript{nd} iter. &   3\textsuperscript{rd} iter. & 1\textsuperscript{st} iter. & 2\textsuperscript{nd} iter. & 3\textsuperscript{rd} iter. \\
\midrule
$\Ltim = \TBC(\Lini)$     & 46.4 & 47.3 & 47.7  &  32.1 & 33.0 & 33.2 \\
$\Lcomb = \comb(\Laug,\Ltim)$ & 48.0 & 48.9 & 49.3  &  34.2 & 35.0 & 35.2 \\

\bottomrule
    \end{tabular}
}
\caption{\textbf{Effect of iterative finetuning} on \ns and \sk validation sets.}
\label{tab:rounds} 
    
\end{table*}

\begin{table*}[!t]
\centering
\newcommand\g{\hphantom{0}}
\let\l\llap
\resizebox{0.8\textwidth}{!}{
\begin{tabular}{l c | c c | c c }
\toprule

\multicolumn{2}{c}{} &  \multicolumn{2}{c}{\ns} &  \multicolumn{2}{c}{\sk} \\
 \cmidrule(lr){3-4}  \cmidrule(lr){5-6}
Labeling & \multicolumn{1}{c}{Iter.} & Label coverage \%   &  mIoU\,\%    &  Label coverage  \%   &  mIoU\,\%      \\
\midrule
$\Lini = \BP(\VLM(\I,\prompts))$ & 0 & 75.8 & 39.3 & 15.4 & 25.5 \\
$\Ltim = \TBC(\Lini)$        & 0 & 74.1 & 39.9 &\g6.7 & 26.3 \\
$\Labc = \ABC(\Lini)$        & 0 & 63.7 & 45.1 & 11.9 & 28.1 \\
$\Laug = \TBC(\ABC(\Lini))$  & 0 & 61.6 & 46.1 &\g4.9 & 29.4 \\
$\Lcomb = \comb(\Laug,\Ltim)$ & 0 & 61.9 & 46.6 &\g5.0 & 31.4 \\
$\Lcomb = \comb(\Laug,\Ltim)$ & 1 &\l100.0& 46.8 &\l100.0&  33.6  \\
$\Lcomb = \comb(\Laug,\Ltim)$ & 2 &\l100.0 & 47.5 &\l100.0&  34.4  \\
\bottomrule
    \end{tabular}
}
\caption{\textbf{Quantity and quality of training labels} computed on \ns and \sk training sets. Note that some lidar points are not visible from the cameras and that the VLM does not assign labels to all pixels. Moreover, while the 6 nuScenes cameras observe the scene at 360\textdegree, the single camera of SemanticKITTI only covers about 1/4 of the lidar field of view. This explains the label coverage.
}
\label{tab:qual} 
    
\end{table*}

\paragraph{Iterative finetuning.} The results in Table~\ref{tab:rounds} show that the quality of labels keeps improving after each round of finetuning, whether starting from $\Ltim$ labels or $\Lcomb$. The best results are obtained when starting from $\Lcomb$, 
which shows that our iterative finetuning strategy is able to maintain the initial advantage provided by our most-refined labels. While a single iteration may provide a gain up to +2.4 mIoU pts, further iterating provides diminishing returns. What thus makes sense practically is to only iterate a few times. In \ours, we perform 3 finetuning iterations.

\paragraph{Pseudo-label analysis.}
In this section, we analyze both the quantity and quality the data annotations used \emph{for training}. The results are presented in Table~\ref{tab:qual} for \ns and \sk. We observe that the VLM struggles more (lower mIoU) on \sk than on \ns. Additionally, with each level of label refinement, the number of labeled points decreases, but the quality of the labels improves. Thanks to our iterative finetuning framework, we are able to annotate the entire datasets after the first iteration while maintaining a high annotation quality.

\paragraph{Comparison to SOTA zero-shot semantic segmentation.}

In Table~\ref{table:sota_semantic_new}, we compare  \ours against SOTA annotation-free methods.
For SAL \cite{sal}, no code is available and the only reported metric is the mIoU on the panoptic segmentation benchmarks, which happens to slightly differ from the mIoU on the semantic segmentation benchmark in the case of \ns (because of slightly different ground truths). We thus report both mIoU\sem\ and mIoU\pan, for the nuScenes semantic and panoptic benchmarks, respectively.

\ours outperforms the SOTA by +3.2 mIoU pts on \ns. It even outperforms methods that additionally use images at inference time by +1.4 mIoU point. The gain over the SOTA on \sk is +6.5 mIoU points (but only a couple of other methods evaluate on this dataset).
\begin{table*}[!t]
\centering
\setlength\extrarowheight{-0.5pt}
\resizebox{0.8\textwidth}{!}{
\begin{tabular}{l c | c c c c c c | c c c c c c }
\toprule

\multicolumn{2}{l}{}  &  \multicolumn{6}{c}{\ns} &  \multicolumn{6}{c}{\sk} \\
 \cmidrule(lr){3-8}  \cmidrule(lr){9-14}
\multicolumn{2}{l}{Method} & PQ  &  RQ  &  SQ   &  PQ$^\text{Th}$  & \multicolumn{1}{c}{PQ$^\text{St}$}  &  mIoU\pan & PQ  &  RQ  &  SQ    &  PQ$^\text{Th}$  &  PQ$^\text{St}$  & mIoU \\

 \midrule

 SAL & \cite{sal} &  38.4 &  47.8 &  \textbf{77.2} &  \textbf{47.5} &  29.2& 33.9 &  24.8 & 32.3 & 66.8 & 17.4 & 30.2 & 28.7 \\ 
SAL-4D &\cite{sal4d} & - & - & - & - &  -  & - & {30.8} &  - &  \textbf{76.9} & 25.5  & \textbf{34.6} & - \\ 

\ours & \clap{(ours)} & \textbf{48.4} & \textbf{58.1} & 71.1  & 46.7 & \textbf{51.3}     & \textbf{49.8}
& \textbf{32.4} & \textbf{41.4}  & 51.8 & \textbf{36.1} & 29.7 & \textbf{35.2} \\
\bottomrule
    \end{tabular}
}
\caption{\textbf{Panoptic segmentation results on \ns and \sk validation sets.}}
\label{tab:panoptic} 
    
\end{table*}

Note that we could not compare to LeAP \cite{leap}, that evaluates with a specific protocol that is not fully described. In particular, the class mapping from the original and classical 19 classes of SemanticKITTI to their 11 classes is ambiguous and not provided in \cite{leap}, while the code is not available. 

Compared to some other approaches, we obtain those SOTA results without resorting to any prompt engineering, VLM tweaking, intricate architecture, complex training, or use of images for inference. Besides, we only have very few and robust parameters, which are the same for all datasets. For instance, we set $\tau$ to 1/3 but any value between 1/4 and 1/2 would have given equal or similar results for both nuScenes and SemanticKITTI. Beyond these bounds, performance would start dropping a~bit.

\begin{figure*}
\centering
\begin{tabular}{ccc}
& Ground truth & \ours \\
%
%
\toprule
    \rotatebox{90}{\hspace{1.5cm}\ns}
&    \includegraphics[width=0.43\linewidth,keepaspectratio] {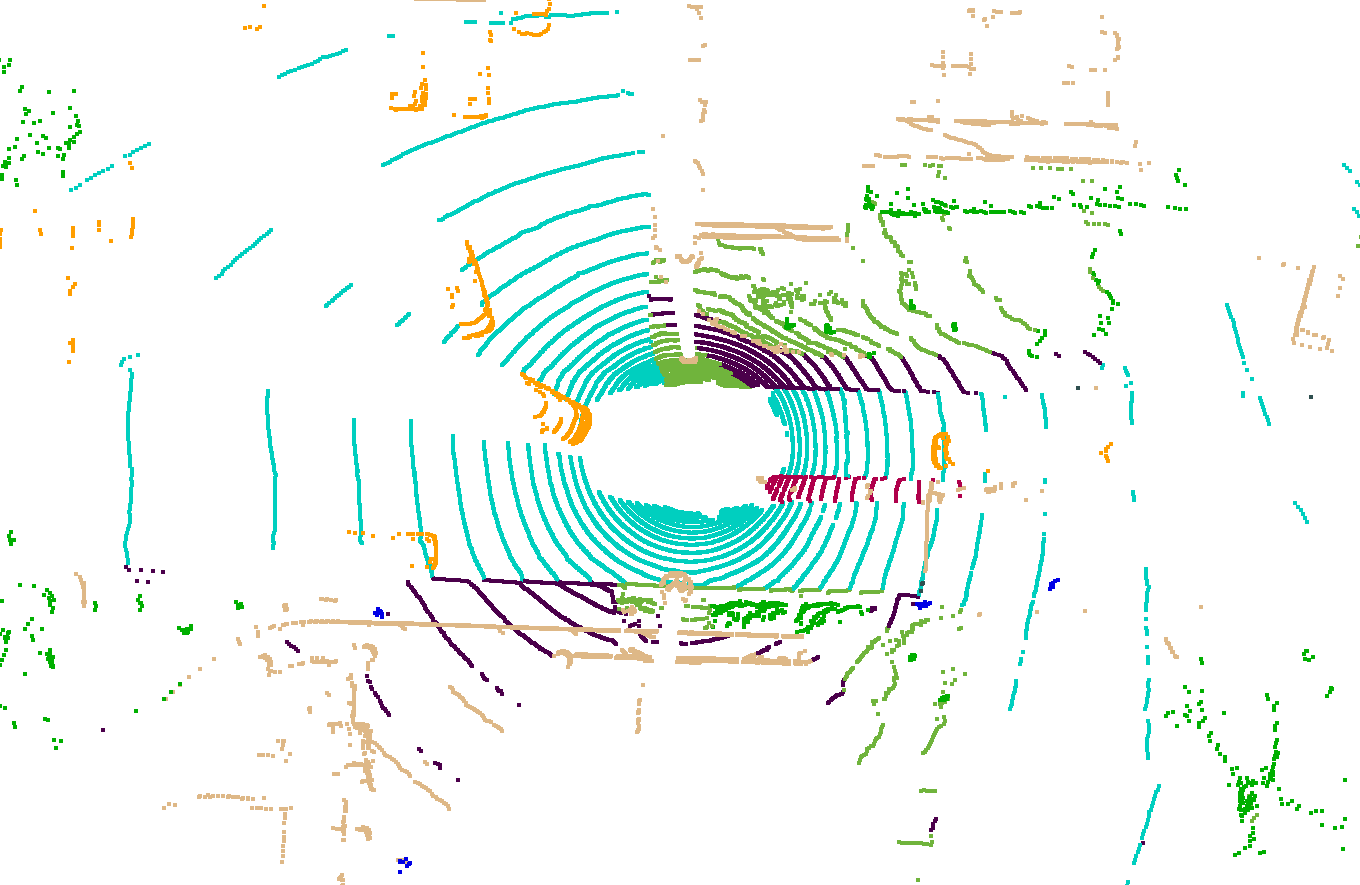}
&    \includegraphics[width=0.43\linewidth,keepaspectratio] {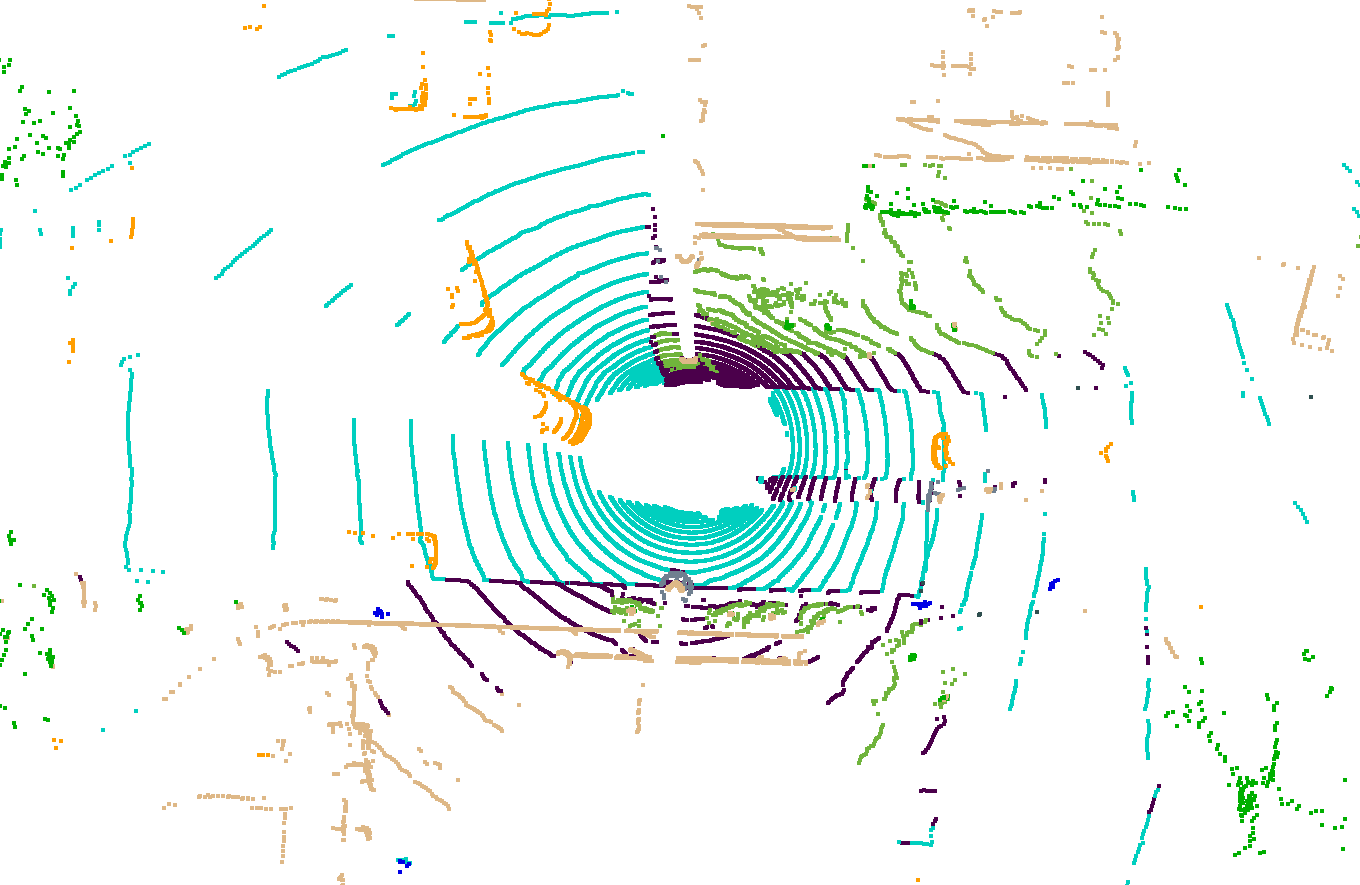}
\\ \hline
    \rotatebox{90}{\hspace{1.5cm}\ns}
&    \includegraphics[width=0.43\linewidth,keepaspectratio] {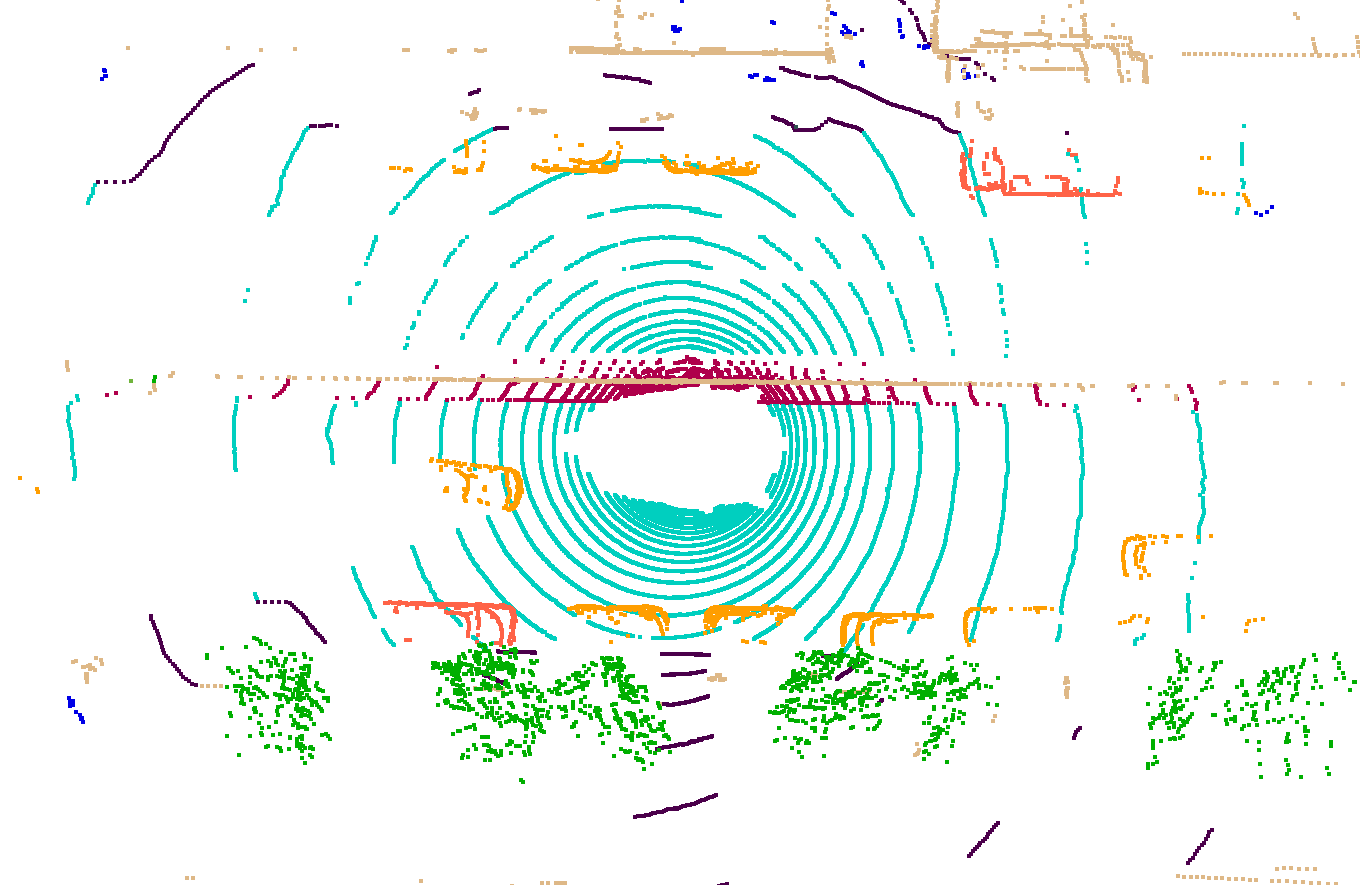} 
&    \includegraphics[width=0.43\linewidth,keepaspectratio] {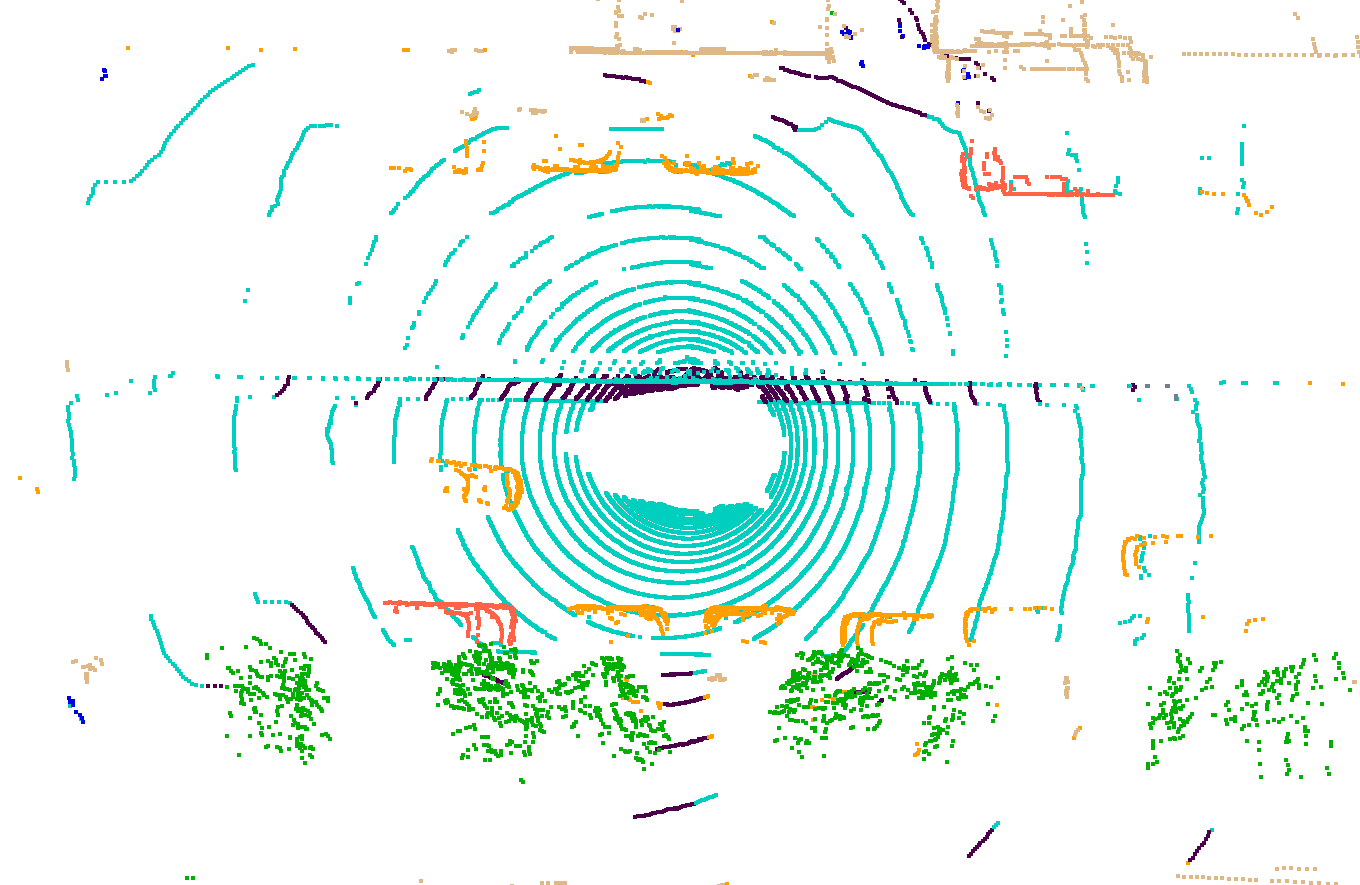}
\\
\midrule
%
%
    \rotatebox{90}{\hspace{1cm}\sk}
&   \includegraphics[width=0.43\linewidth,keepaspectratio] {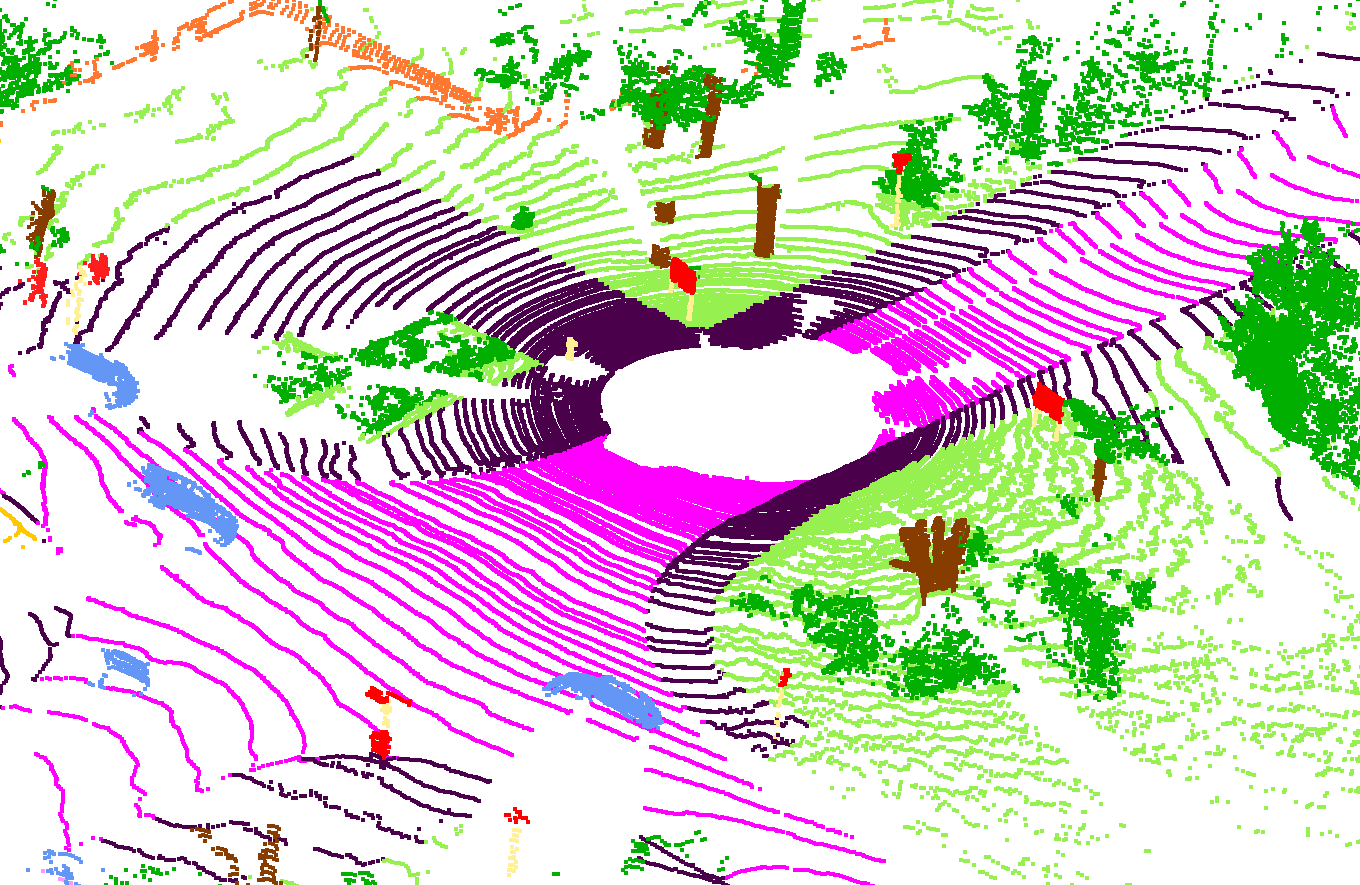}
&
\includegraphics[width=0.43\linewidth,keepaspectratio] {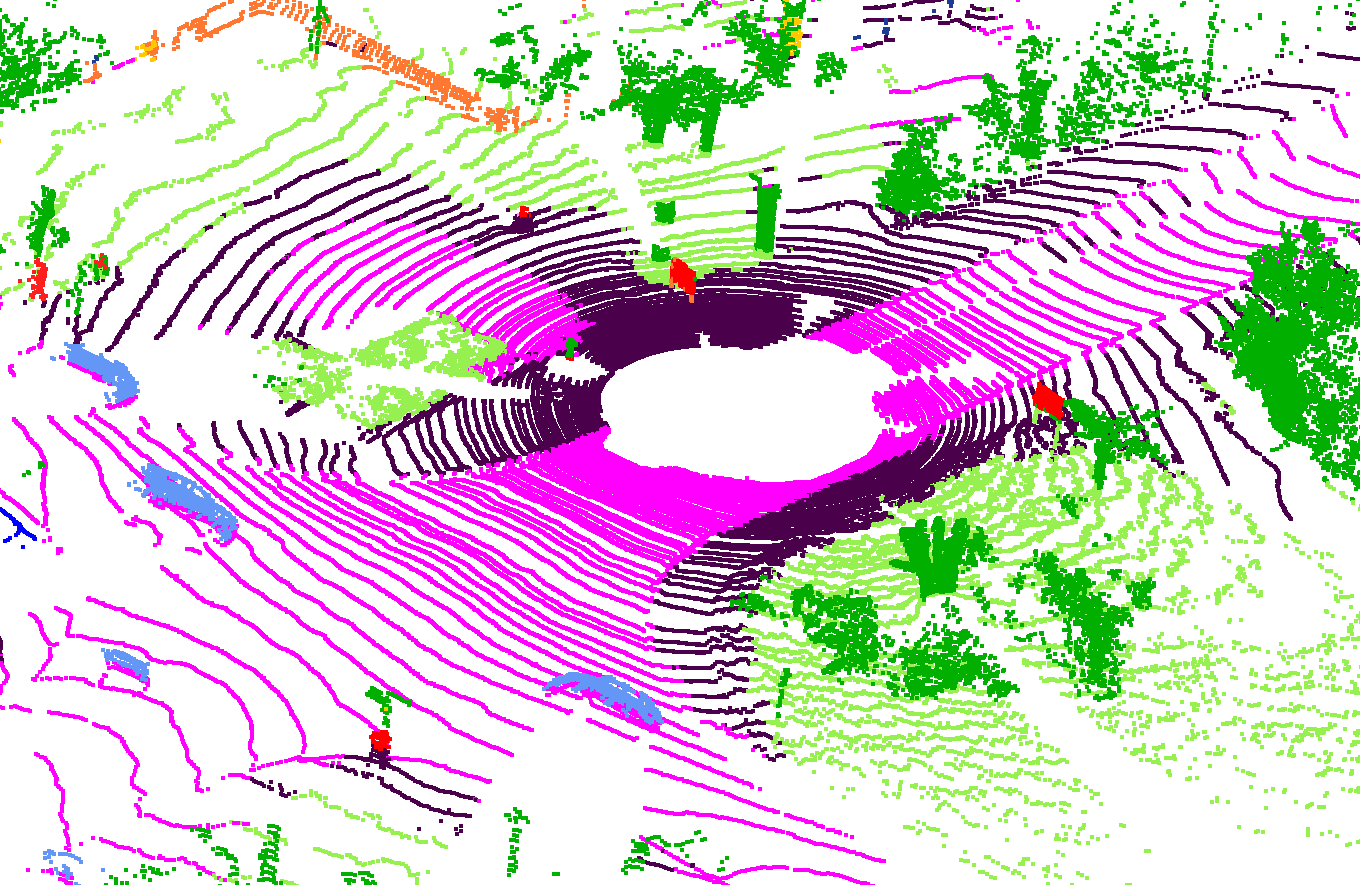}
\\ \hline
    \rotatebox{90}{\hspace{1cm}\sk}
&    \includegraphics[width=0.43\linewidth,keepaspectratio] {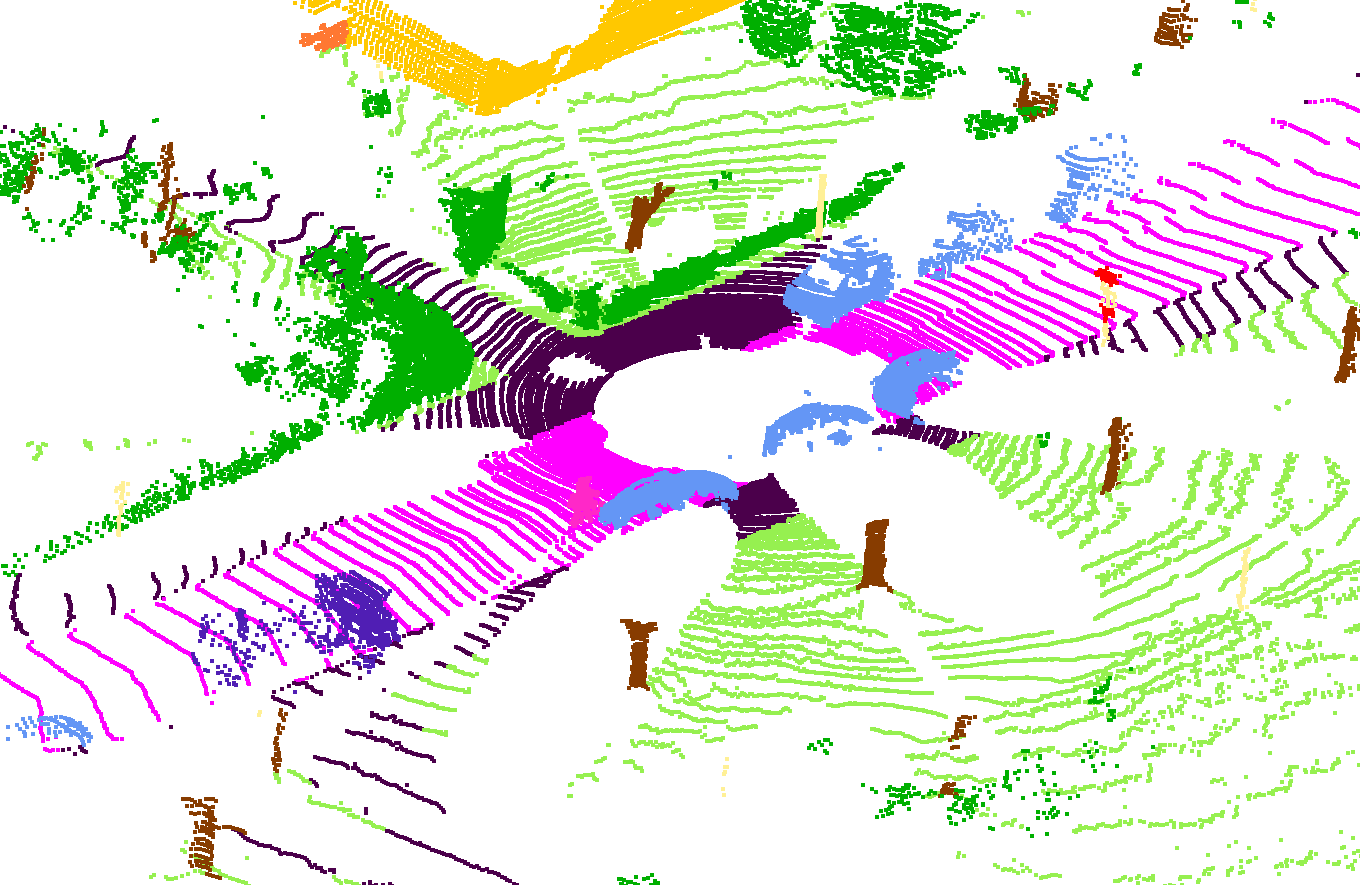}
&    \includegraphics[width=0.43\linewidth,keepaspectratio] {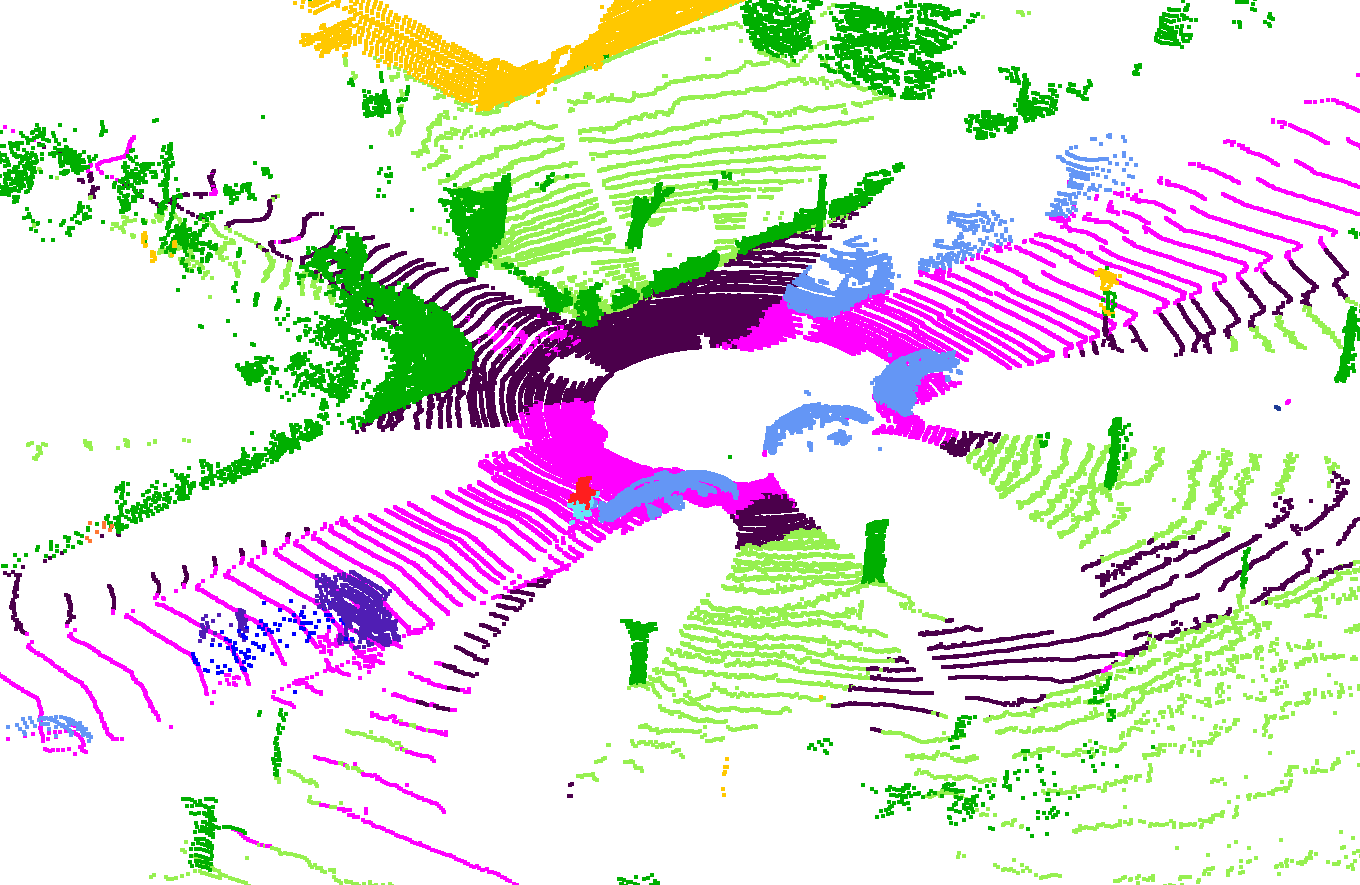}
\\
\bottomrule
\end{tabular}
\caption{Qualitative results of semantic segmentation from the validation sets of \ns and \sk. The color code used to represent each class is provided in Supplementary Material.
A typical error with \ours on both datasets is a confusion between different types of flat surfaces such as \relax{\emph{road/driveable surface}, \emph{sidewalk} and \emph{terrain}. In \sk, \emph{trunks} are also systematically included in \emph{vegetation} rather than considered as a separate class. These observations are consistent with the quantitative class-wise results provided in Supplementary Material.}}

\label{fig:visuals}
\end{figure*}

We present in Figure~\ref{fig:visuals} qualitative results on the \ns and \sk validation sets. 

\begin{table}[t]
\begin{center}

\resizebox{0.48\textwidth}{!}{

\begin{tabular}{l@{~}c|c|c}
\toprule

\multicolumn{2}{l|}{Method}
    & {\ns}

    & {\sk}

\\
\midrule
\rowcolor{gray!40}    \wi (full sup.)& \cite{waffleiron}
    & 78.7
    & 63.4

\\  
\midrule

CLIP2Scene & \cite{clip2scene}
    & 20.8    
    & -

\\  
Towards\,VFMs & \cite{chen2023cns}
    & 26.8
    
    & -

\\  
AutoVoc3D w.\ LAVE & \cite{autovoc}
    & 30.6

    & -

\\
AdaCo & \cite{zou2025adaco}
    & 31.2

    & 25.7

\\

SAL & \cite{sal}

    & 33.9\rlap{ \textsuperscript{\textdagger}}
    & 28.7
\\

OV3D w/ OpenScene-3D & \cite{jiang2024ov3d}
    & \relax{45.5}

    & -

\\
GGSD & \cite{wang2024ggsd}
    & \relax{46.1}

    & -

\\

\ours & \clap{(ours)}
    & \textbf{49.3}    
    &\textbf{35.2}

\\
\midrule
\rlap{\emph{Method additionally using images at inference time}}
\\
OpenScene - OpenSeg & \cite{openscene}
    & 42.1 
    
    & -

\\
AFOV & \cite{sunaaai25}
    & 47.9 
    & -

\\

\bottomrule
\end{tabular}
}
\end{center} \vspace{-5mm}
\caption{\textbf{Semantic segmentation results on \ns and \sk validation sets.}
$\dagger$~indicates the panoptic mIoU, which, in the case of nuScenes, slightly differs ($<$~1 pt in general) from the classical semantic segmentation mIoU otherwise reported here (see appendix).
}
\label{table:sota_semantic_new}
\end{table}

\paragraph{Comparison to SOTA zero-shot panoptic segmentation.}

We further evaluate our method on the panoptic segmentation task, comparing it against previous annotation-free approaches in Table~\ref{tab:panoptic} for both \ns and \sk.

To obtain our panoptic segments, we apply ALPINE~\cite{alpine} on top of our semantic segmentation predictions. Given the semantic segmentation maps, ALPINE projects points into the BEV space, performs clustering by building a kNN graph and extracts connected components. The method is learning-free. 
Our method achieves a significant improvement on the main panoptic metric, outperforming SAL by +10 PQ points or more on both datasets.
\ours also outperforms SAL-4D on \sk by nearly 2 PQ points, with particularly significant improvement observed in instance segmentation (PQ$^\text{Th}$). 

\paragraph{Moving objects.}
\label{sec:limits}
The spatio-temporal consolidation (\TBC) assumes that voxel semantics remains constant over time. This is true for the static part of the scene, which represents most of the points. However, it may not be true for moving objects. Yet, there is no issue when an empty area is traversed by a moving object: in corresponding voxels, there are no points before the traversal, some points during the traversal, and again no points afterwards. Therefore, voxel-based voting still makes sense. 

An actual issue occurs only when two different objects traverse the same empty space, and then only in one particular case: if the objects belong to different classes (e.g., car and bicycle) and if the object speed and frame rate are such that points from different objects are actually scanned within the same voxel. Given the distribution of moving object classes (a vast majority of cars), the low frame rate (2\,Hz for nuScenes, 10\,Hz for SemanticKITTI), and the small voxel size (10\,cm), this happens rarely.

In other words, while dynamic environments can \emph{in theory} break the voxel semantic consistency, they rarely do so \emph{in practice}. The fact is, \ours outperforms all other methods while the evaluation datasets do include moving objects. Moreover, if needed, time consistency for moving objects can be improved by cutting sequences into smaller sections.

\section{Conclusion}
\label{sec:conc}

Our work shows it is possible to transfer the image segmentation capability of an off-the-shelf open-vocabulary 2D-based VLM, used in a black box manner, in order to segment 3D lidar scans and largely outperform the SOTA. Besides, compared to the bells and whistles of existing approaches, our method is relatively simple and relies on very few and little-sensitive parameters.

Our approach is not bound to 2D VLMs. It could be also applied to any strong 2D semantic segmenter trained with full-supervision, alleviating the need for 3D annotation.

\paragraph*{Acknowledgments.}
We acknowledge EuroHPC Joint Undertaking for awarding the project ID EHPC-REG-2024R02-234 access to Karolina, Czech Republic. This work was granted access to the HPC resources of IDRIS under the allocations AD011014946R1 made by GENCI. 
This work was also supported by the European Union’s Horizon Europe research and innovation programme under grant agreement No 101214398 (ELLIOT).

{
    \bibliographystyle{ieeenat_fullname}
    \bibliography{main}  

\begin{thebibliography}{40}
\providecommand{\natexlab}[1]{#1}
\providecommand{\url}[1]{\texttt{#1}}
\expandafter\ifx\csname urlstyle\endcsname\relax
  \providecommand{\doi}[1]{doi: #1}\else
  \providecommand{\doi}{doi: \begingroup \urlstyle{rm}\Url}\fi

\bibitem[Akter et~al.(2025)Akter, Shihab, and Sharma]{akter2025imagesegmentationllmsurvey}
Sanjeda Akter, Ibne~Farabi Shihab, and Anuj Sharma.
\newblock Image segmentation with large language models: A survey with perspectives for intelligent transportation systems, 2025.
\newblock arXiv preprint 2506.14096.

\bibitem[Behley et~al.(2019)Behley, Garbade, Milioto, Quenzel, Behnke, Stachniss, and Gall]{semantickitti}
Jens Behley, Martin Garbade, Andres Milioto, Jan Quenzel, Sven Behnke, Cyrill Stachniss, and Jurgen Gall.
\newblock {{SemanticKITTI}: A Dataset for Semantic Scene Understanding of LiDAR Sequences}.
\newblock In \emph{ICCV}, pages 9297--9307, 2019.

\bibitem[Buslaev et~al.(2018)Buslaev, Parinov, Khvedchenya, Iglovikov, and Kalinin]{albumentations}
A. Buslaev, A. Parinov, E. Khvedchenya, V.~I. Iglovikov, and A.~A. Kalinin.
\newblock {Albumentations: fast and flexible image augmentations}.
\newblock \emph{ArXiv e-prints}, 2018.

\bibitem[Caesar et~al.(2020)Caesar, Bankiti, Lang, Vora, Liong, Xu, Krishnan, Pan, Baldan, and Beijbom]{nuscenes}
Holger Caesar, Varun Bankiti, Alex~H. Lang, Sourabh Vora, Venice~Erin Liong, Qiang Xu, Anush Krishnan, Yu Pan, Giancarlo Baldan, and Oscar Beijbom.
\newblock {nuScenes}: A multimodal dataset for autonomous driving.
\newblock In \emph{CVPR}, 2020.

\bibitem[Chen et~al.(2023{\natexlab{a}})Chen, Liu, Kong, Chen, Xinge, Ma, Liu, and Wang]{chen2023towards}
Runnan Chen, Youquan Liu, Lingdong Kong, Nenglun Chen, ZHU Xinge, Yuexin Ma, Tongliang Liu, and Wenping Wang.
\newblock Towards label-free scene understanding by vision foundation models.
\newblock In \emph{NeurIPS}, 2023{\natexlab{a}}.

\bibitem[Chen et~al.(2023{\natexlab{b}})Chen, Liu, Kong, Chen, ZHU, Ma, Liu, and Wang]{chen2023cns}
Runnan Chen, Youquan Liu, Lingdong Kong, Nenglun Chen, Xinge ZHU, Yuexin Ma, Tongliang Liu, and Wenping Wang.
\newblock Towards label-free scene understanding by vision foundation models.
\newblock In \emph{Advances in Neural Information Processing Systems (NeurIPS)}, 2023{\natexlab{b}}.

\bibitem[Chen et~al.(2023{\natexlab{c}})Chen, Liu, Kong, Zhu, Ma, Li, Hou, Qiao, and Wang]{clip2scene}
Runnan Chen, Youquan Liu, Lingdong Kong, Xinge Zhu, Yuexin Ma, Yikang Li, Yuenan Hou, Yu Qiao, and Wenping Wang.
\newblock Clip2scene: Towards label-efficient 3d scene understanding by clip.
\newblock In \emph{CVPR}, 2023{\natexlab{c}}.

\bibitem[Chen et~al.(2025)Chen, Huang, Fan, Wang, Zhou, Guan, and Lin]{chen2025reproducible}
Ziliang Chen, Xin Huang, Xiaoxuan Fan, Keze Wang, Yuyu Zhou, Quanlong Guan, and Liang Lin.
\newblock Reproducible vision-language models meet concepts out of pre-training.
\newblock In \emph{CVPR}, 2025.

\bibitem[Ding et~al.(2023)Ding, Wang, and Tu]{ding2023maskclip}
Zheng Ding, Jieke Wang, and Zhuowen Tu.
\newblock Open-vocabulary universal image segmentation with {MaskCLIP}.
\newblock In \emph{ICML}, 2023.

\bibitem[Feng et~al.(2025)Feng, Liu, Yang, Cai, Zhang, Zhan, Huang, Yan, Wan, Liu, Wang, Lv, Liu, Shi, Liu, and Wang]{feng2025vlmobjectdetection}
Yongchao Feng, Yajie Liu, Shuai Yang, Wenrui Cai, Jinqing Zhang, Qiqi Zhan, Ziyue Huang, Hongxi Yan, Qiao Wan, Chenguang Liu, Junzhe Wang, Jiahui Lv, Ziqi Liu, Tengyuan Shi, Qingjie Liu, and Yunhong Wang.
\newblock Vision-language model for object detection and segmentation: A review and evaluation, 2025.
\newblock arXiv preprint 2504.09480.

\bibitem[Gebraad et~al.(2025)Gebraad, Palffy, and Caesar]{leap}
Simon Gebraad, Andras Palffy, and Holger Caesar.
\newblock Leap: Consistent multi-domain 3d labeling using foundation models.
\newblock \emph{arXiv preprint arXiv:2502.03901}, 2025.

\bibitem[Ghiasi et~al.(2022)Ghiasi, Gu, Cui, and Lin]{openseg}
Golnaz Ghiasi, Xiuye Gu, Yin Cui, and Tsung-Yi Lin.
\newblock Scaling open-vocabulary image segmentation with image-level labels.
\newblock In \emph{ECCV}, 2022.

\bibitem[Jia et~al.(2021)Jia, Yang, Xia, Chen, Parekh, Pham, Le, Sung, Li, and Duerig]{jia2021align}
Chao Jia, Yinfei Yang, Ye Xia, Yi{-}Ting Chen, Zarana Parekh, Hieu Pham, Quoc~V. Le, Yun{-}Hsuan Sung, Zhen Li, and Tom Duerig.
\newblock Scaling up visual and vision-language representation learning with noisy text supervision.
\newblock In \emph{ICML}, 2021.

\bibitem[Jiang et~al.(2024)Jiang, Shi, and Schiele]{jiang2024ov3d}
Li Jiang, Shaoshuai Shi, and Bernt Schiele.
\newblock Open-vocabulary {3D} semantic segmentation with foundation models.
\newblock In \emph{CVPR}, 2024.

\bibitem[Kirillov et~al.(2023)Kirillov, Mintun, Ravi, Mao, Rolland, Gustafson, Xiao, Whitehead, Berg, Lo, Doll{\'a}r, and Girshick]{sam}
Alexander Kirillov, Eric Mintun, Nikhila Ravi, Hanzi Mao, Chloe Rolland, Laura Gustafson, Tete Xiao, Spencer Whitehead, Alexander~C. Berg, Wan-Yen Lo, Piotr Doll{\'a}r, and Ross Girshick.
\newblock Segment anything.
\newblock In \emph{ICCV}, 2023.

\bibitem[Li et~al.(2022)Li, Weinberger, Belongie, Koltun, and Ranftl]{li2022lseg}
Boyi Li, Kilian~Q Weinberger, Serge Belongie, Vladlen Koltun, and Rene Ranftl.
\newblock Language-driven semantic segmentation.
\newblock In \emph{ICLR}, 2022.

\bibitem[Li et~al.(2025)Li, Wu, Du, Nghiem, and Shi]{li2025surveystateartlarge}
Zongxia Li, Xiyang Wu, Hongyang Du, Huy Nghiem, and Guangyao Shi.
\newblock A survey of state of the art large vision language models: Alignment, benchmark, evaluations and challenges.
\newblock \emph{arXiv:2501.02189}, 2025.

\bibitem[Ma et~al.(2024)Ma, Bhalgat, Smart, Chen, Li, Ding, Gu, Chen, Peng, Bian, Torr, Pollefeys, Nießner, Reid, Chang, Laina, and Prisacariu]{ma2024llmsstep3dworld}
Xianzheng Ma, Yash Bhalgat, Brandon Smart, Shuai Chen, Xinghui Li, Jian Ding, Jindong Gu, Dave~Zhenyu Chen, Songyou Peng, Jia-Wang Bian, Philip~H Torr, Marc Pollefeys, Matthias Nießner, Ian~D Reid, Angel~X. Chang, Iro Laina, and Victor~Adrian Prisacariu.
\newblock When {LLMs} step into the 3d world: A survey and meta-analysis of {3D} tasks via multi-modal large language models, 2024.
\newblock arXiv prepreint 2405.10255.

\bibitem[O{\v{s}}ep et~al.(2024)O{\v{s}}ep, Meinhardt, Ferroni, Peri, Ramanan, and Leal-Taix{\'e}]{sal}
Aljo{\v{s}}a O{\v{s}}ep, Tim Meinhardt, Francesco Ferroni, Neehar Peri, Deva Ramanan, and Laura Leal-Taix{\'e}.
\newblock Better call sal: Towards learning to segment anything in lidar.
\newblock In \emph{European Conference on Computer Vision}, pages 71--90. Springer, 2024.

\bibitem[Peng et~al.(2023)Peng, Genova, Jiang, Tagliasacchi, Pollefeys, Funkhouser, et~al.]{openscene}
Songyou Peng, Kyle Genova, Chiyu Jiang, Andrea Tagliasacchi, Marc Pollefeys, Thomas Funkhouser, et~al.
\newblock Openscene: 3d scene understanding with open vocabularies.
\newblock In \emph{Proceedings of the IEEE/CVF conference on computer vision and pattern recognition}, pages 815--824, 2023.

\bibitem[Puy et~al.(2023)Puy, Boulch, and Marlet]{waffleiron}
Gilles Puy, Alexandre Boulch, and Renaud Marlet.
\newblock Using a waffle iron for automotive point cloud semantic segmentation.
\newblock In \emph{ICCV}, 2023.

\bibitem[Puy et~al.(2024)Puy, Gidaris, Boulch, Siméoni, Sautier, Pérez, Bursuc, and Marlet]{scalr}
Gilles Puy, Spyros Gidaris, Alexandre Boulch, Oriane Siméoni, Corentin Sautier, Patrick Pérez, Andrei Bursuc, and Renaud Marlet.
\newblock Three pillars improving vision foundation model distillation for lidar.
\newblock In \emph{CVPR}, 2024.

\bibitem[Radford et~al.(2021)Radford, Kim, Hallacy, Ramesh, Goh, Agarwal, Sastry, Askell, Mishkin, Clark, et~al.]{radford2021clip}
Alec Radford, Jong~Wook Kim, Chris Hallacy, Aditya Ramesh, Gabriel Goh, Sandhini Agarwal, Girish Sastry, Amanda Askell, Pamela Mishkin, Jack Clark, et~al.
\newblock Learning transferable visual models from natural language supervision.
\newblock In \emph{ICML}, 2021.

\bibitem[Ren et~al.(2024)Ren, Liu, Zeng, Lin, Li, Cao, Chen, Huang, Chen, Yan, Zeng, Zhang, Li, Yang, Li, Jiang, and Zhang]{gsam}
Tianhe Ren, Shilong Liu, Ailing Zeng, Jing Lin, Kunchang Li, He Cao, Jiayu Chen, Xinyu Huang, Yukang Chen, Feng Yan, Zhaoyang Zeng, Hao Zhang, Feng Li, Jie Yang, Hongyang Li, Qing Jiang, and Lei Zhang.
\newblock Grounded sam: Assembling open-world models for diverse visual tasks, 2024.

\bibitem[Sautier et~al.(2025)Sautier, Puy, Boulch, Marlet, and Lepetit]{alpine}
Corentin Sautier, Gilles Puy, Alexandre Boulch, Renaud Marlet, and Vincent Lepetit.
\newblock Clustering is back: Reaching state-of-the-art {LiDAR} instance segmentation without training.
\newblock \emph{arxiv}, 2025.

\bibitem[Schuhmann et~al.(2022)Schuhmann, Beaumont, Vencu, Gordon, Wightman, Cherti, Coombes, Katta, Mullis, Wortsman, Schramowski, Kundurthy, Crowson, Schmidt, Kaczmarczyk, and Jitsev]{schuhmann2002laion5b}
Christoph Schuhmann, Romain Beaumont, Richard Vencu, Cade Gordon, Ross Wightman, Mehdi Cherti, Theo Coombes, Aarush Katta, Clayton Mullis, Mitchell Wortsman, Patrick Schramowski, Srivatsa Kundurthy, Katherine Crowson, Ludwig Schmidt, Robert Kaczmarczyk, and Jenia Jitsev.
\newblock {LAION-5B}: an open large-scale dataset for training next generation image-text models.
\newblock In \emph{NeurIPS}, 2022.

\bibitem[Sun et~al.(2025)Sun, Liu, Wang, Tian, Chen, and Wang]{sunaaai25}
Boyi Sun, Yuhang Liu, Xingxia Wang, Bin Tian, Long Chen, and Fei-Yue Wang.
\newblock 3d annotation-free learning by distilling 2d open-vocabulary segmentation models for autonomous driving.
\newblock In \emph{AAAI}, 2025.

\bibitem[Wang et~al.(2024)Wang, Wang, Li, Zhang, Lei, and Zhang]{wang2024ggsd}
Pengfei Wang, Yuxi Wang, Shuai Li, Zhaoxiang Zhang, Zhen Lei, and Lei Zhang.
\newblock Open vocabulary 3d scene understanding via geometry guided self-distillation.
\newblock In \emph{ECCV}, 2024.

\bibitem[Wei et~al.(2024)Wei, {\"U}lger, Nejadasl, Gevers, and Oswald]{autovoc}
Weijie Wei, Osman {\"U}lger, Fatemeh~Karimi Nejadasl, Theo Gevers, and Martin~R Oswald.
\newblock Auto-vocabulary segmentation for lidar points.
\newblock \emph{arXiv preprint arXiv:2406.09126}, 2024.

\bibitem[Wu et~al.(2024{\natexlab{a}})Wu, Li, Xu, Yuan, Ding, Yang, Li, Zhang, Tong, Jiang, Ghanem, and Tao]{wu2024openvocsurvey}
Jianzong Wu, Xiangtai Li, Shilin Xu, Haobo Yuan, Henghui Ding, Yibo Yang, Xia Li, Jiangning Zhang, Yunhai Tong, Xudong Jiang, Bernard Ghanem, and Dacheng Tao.
\newblock Towards open vocabulary learning: A survey.
\newblock \emph{TPAMI}, 46\penalty0 (7):\penalty0 5092–5113, 2024{\natexlab{a}}.

\bibitem[Wu et~al.(2024{\natexlab{b}})Wu, Jiang, Wang, Liu, Liu, Qiao, Ouyang, He, and Zhao]{wu2024ptv3}
Xiaoyang Wu, Li Jiang, Peng-Shuai Wang, Zhijian Liu, Xihui Liu, Yu Qiao, Wanli Ouyang, Tong He, and Hengshuang Zhao.
\newblock Point transformer v3: Simpler, faster, stronger.
\newblock In \emph{CVPR}, 2024{\natexlab{b}}.

\bibitem[Wysoczanska et~al.(2024)Wysoczanska, Ramamonjisoa, Trzcinski, and Sim\'eoni]{clipdiy}
Monika Wysoczanska, Michael Ramamonjisoa, Tomasz Trzcinski, and Oriane Sim\'eoni.
\newblock Clip-diy: Clip dense inference yields open-vocabulary semantic segmentation for-free.
\newblock In \emph{WACV}, 2024.

\bibitem[Zhang et~al.(2023)Zhang, Li, Zou, Liu, Li, Yang, and Zhang]{openseed}
Hao Zhang, Feng Li, Xueyan Zou, Shilong Liu, Chunyuan Li, Jianwei Yang, and Lei Zhang.
\newblock A simple framework for open-vocabulary segmentation and detection.
\newblock In \emph{ICCV}, 2023.

\bibitem[Zhang et~al.(2024)Zhang, Huang, Jin, and Lu]{zhang2024vlmsurvey}
Jingyi Zhang, Jiaxing Huang, Sheng Jin, and Shijian Lu.
\newblock Vision-language models for vision tasks: A survey.
\newblock \emph{TPAMI}, 46\penalty0 (8):\penalty0 5625--5644, 2024.

\bibitem[Zhang et~al.(2025)Zhang, O{\v{s}}ep, Leal-Taix{\'e}, and Meinhardt]{sal4d}
Yushan Zhang, Aljo{\v{s}}a O{\v{s}}ep, Laura Leal-Taix{\'e}, and Tim Meinhardt.
\newblock Zero-shot 4d lidar panoptic segmentation.
\newblock \emph{arXiv preprint arXiv:2504.00848}, 2025.

\bibitem[Zhong et~al.(2022)Zhong, Yang, Zhang, Li, Codella, Li, Zhou, Dai, Yuan, Li, and Gao]{regionclip}
Yiwu Zhong, Jianwei Yang, Pengchuan Zhang, Chunyuan Li, Noel Codella, Liunian~Harold Li, Luowei Zhou, Xiyang Dai, Lu Yuan, Yin Li, and Jianfeng Gao.
\newblock {RegionCLIP}: Region-based language-image pretraining.
\newblock In \emph{CVPR}, 2022.

\bibitem[Zhou et~al.(2022)Zhou, Change~Loy, and Dai]{zhou2022maskclip}
Chong Zhou, Chen Change~Loy, and Bo Dai.
\newblock Extract free dense labels from clip.
\newblock In \emph{ECCV}, 2022.

\bibitem[Zhou et~al.(2024)Zhou, Xia, Zhang, Chang, Wang, Yuan, Konukoglu, and Cremers]{zhou2024imagesegmentationfoundationmodel}
Tianfei Zhou, Wang Xia, Fei Zhang, Boyu Chang, Wenguan Wang, Ye Yuan, Ender Konukoglu, and Daniel Cremers.
\newblock Image segmentation in foundation model era: A survey.
\newblock \emph{arXiv:2408.12957}, 2024.

\bibitem[Zou et~al.(2025)Zou, Zhao, Huang, Xia, Wen, Li, and Wang]{zou2025adaco}
Pufan Zou, Shijia Zhao, Weijie Huang, Qiming Xia, Chenglu Wen, Wei Li, and Cheng Wang.
\newblock {AdaCo}: Overcoming visual foundation model noise in {3D} semantic segmentation via adaptive label correction.
\newblock In \emph{AAAI}, 2025.

\bibitem[Zou et~al.(2023)Zou, Yang, Zhang, Li, Li, Wang, Wang, Gao, and Lee]{seem}
Xueyan Zou, Jianwei Yang, Hao Zhang, Feng Li, Linjie Li, Jianfeng Wang, Lijuan Wang, Jianfeng Gao, and Yong~Jae Lee.
\newblock Segment everything everywhere all at once.
\newblock \emph{Advances in neural information processing systems}, 36:\penalty0 19769--19782, 2023.

\end{thebibliography}
}
\newpage
\appendix
\section{Appendix}
We supplement the main paper with the performance on super-classes (\cref{sec:sc-results}), classwise results (\cref{sec:classwise-results}), information about the quantity and quality of labeled points for training (\cref{sec:labeled_points}), details about prompts used in experiments (\cref{sec:prompts}), and a detailed study concerning the use of augmentations (\cref{sec:augmentation}).

\section{Results with super-classes}
\label{sec:sc-results}

Some methods like SAL \cite{sal} and LeAP \cite{leap} evaluate on broader super-classes. For instance, SAL merges car, motorcycle, truck, etc., into a single `vehicle' class. In \cref{table:sc_semantic}, we recall results with the original fine-grained classes and also report our results adopting the same merging strategy as in SAL~\cite{sal}, applying class merging during inference to predict super-class labels. (As said in the experiment section of the paper, no comparison is possible with LeAP, which does not provides its mapping from 19 classes into 11.)

Besides, no code is available for SAL and the only reported metric is the mIoU on the panoptic segmentation benchmarks, which happens to slightly differ ($<$~1 pt in practice) from the mIoU on the semantic segmentation benchmark in the specific case of \ns, because of slightly different ground truths in both benchmarks. We thus report in \cref{table:sc_semantic} both mIoU\sem\ and mIoU\pan, for the nuScenes semantic and panoptic benchmarks, respectively. As can be seen in the table, \ours clearly outperforms SAL on semantic segmentation, with both fine-grained classes and super-classes.

\begin{table*}[h!]
\begin{center}

\begin{tabular}{l@{~}c|c@{}c c@{~~}c|c@{~~}c}
\toprule
\multicolumn{2}{l}{Method}
    & mIoU\sem
    & \multicolumn{1}{c}{mIoU\pan}
    & \cellcolor{blue!17}\textit{mIoU\semsc}
    & \cellcolor{blue!17}\textit{mIoU\pansc}
    & mIoU
    & \cellcolor{blue!17}\textit{mIoU\sc}
\\
\midrule
SAL & \cite{sal}
    & -
    & 33.9
    & \cellcolor{blue!17}-
    & \cellcolor{blue!17}\textit{52.6}
    
    & 28.7
    & \cellcolor{blue!17}\textit{52.8}
\\  
\ours & (ours)
    & {49.3}
    & \textbf{49.8}
    
    & \cellcolor{blue!17}\textit{68.4}
    & \cellcolor{blue!17}-
    
    & \textbf{35.2}
    & \cellcolor{blue!17}\textit{\textbf{67.4}}
\\
\bottomrule
\end{tabular}
\end{center}
\caption{\textbf{Semantic segmentation results on \ns and \sk validation sets for both original fine-grained classes and super-classes.}
For \ns, mIoU\sem\ is computed with semantic segmentation ground truth while mIoU\pan\ is computed with the panoptic segmentation ground truth. 
mIoU\sc, in \smash{\colorbox{blue!17}{blue}}, corresponds to a mapping of all classes into six ``super-classes'' (sc), as defined in SAL \cite{sal}.}
\label{table:sc_semantic}
\end{table*}

\section{Classwise results}
\label{sec:classwise-results}

Tables~\ref{table:class_wise_iou_ns} and~\ref{table:class_wise_iou_sk} respectively present classwise semantic segmentation results on \ns \cite{nuscenes} and \sk \cite{semantickitti} val sets, evaluated with networks initially finetuned on labelings $\Ltim$ or $\Lcomb$ (1\textsuperscript{st} iter.), before iterated training (2\textsuperscript{nd} and 3\textsuperscript{rd} iter.).%
\begin{table*}[h!]
\begin{center}
\def\best{\cellcolor{blue!17}}
\begin{tabular}{l|rrr|rrr|r}
\toprule
\textbf{\ns}    & \multicolumn{3}{c|}{$\Ltim$}
    & \multicolumn{3}{c|}{$\Lcomb$}
    & Full\hspace{0.5ex}
\\
 \cmidrule(lr){2-4}  \cmidrule(lr){5-7}
    \textbf{Classes} (default)
    & 1\textsuperscript{st} iter.\hspace{-1ex}
    & 2\textsuperscript{nd} iter.\hspace{-1ex}
    & 3\textsuperscript{rd} iter.\hspace{-1ex}
    & 1\textsuperscript{st} iter.\hspace{-1ex}
    & 2\textsuperscript{nd} iter.\hspace{-1ex}
    & 3\textsuperscript{rd} iter.\hspace{-1ex}
    & supp.\hspace{-0.5ex}
\\
\midrule
barrier    & 2.8
    & 3.2
    & \best{3.6}
    & 4.3
    & 4.0
    & \best{5.2}
    & 79.2
\\ 
bicycle    & \best{28.2}
    & 28.0
    & 27.7
    & 21.2
    & 20.4
    & \best{21.3}
    & 53.2
\\ 
bus    & \best{72.5}
    & 72.0
    & 71.0
    & 80.2
    & \best{82.2}
    & 80.5
    & 92.5
\\ 
car    & 63.7
    & 65.2
    & \best{66.7}
    & 64.4
    & 66.6
    & \best{67.5}
    & 88.1
\\ 
construction vehicle    & 2.4
    & 4.6
    & \best{5.9}
    & 9.3
    & 9.8
    & \best{10.6}
    & 50.4
\\ 
motorcycle    & 57.0
    & 57.6
    & \best{58.7}
    & 62.6
    & 63.5
    & \best{64.1}
    & 87.8
\\ 
pedestrian    & 67.1
    & 67.9
    & \best{68.2}
    & 70.8
    & 72.1
    & \best{72.7}
    & 83.4
\\ 
traffic cone    & 39.7
    & 44.9
    & \best{46.5}
    & 42.3
    & 45.5
    & \best{46.3}
    & 70.3
\\ 
trailer    & 0.0
    & \best{0.1}
    & \best{0.1}
    & \best{0.1}
    & 0.0
    & 0.0
    & 73.5
\\ 
truck    & 57.9
    & 59.5 
    & \best{59.7}
    & 56.5
    & 57.7
    & \best{58.2}
    & 84.1
\\ 
driveable surface    & 89.4
    & \best{89.6}
    & \best{89.6}
    & \best{90.5}
    & 90.4
    & 90.1
    & 96.9
\\ 
other flat    & 0.0
    & 0.0
    & 0.0
    & 0.0
    & 0.0
    & 0.0
    & 73.3
\\ 
sidewalk    & 48.6
    & 49.3
    & \best{49.6}
    & 50.0
    & 50.7
    & \best{51.0}
    & 75.5
\\ 
terrain    & 59.5
    & 60.2
    & \best{60.4}
    & 61.3
    & \best{61.7}
    & 61.5
    & 75.4
\\ 
manmade    & 74.4
    & 75.8
    & \best{76.3}
    & 74.6
    & 76.5
    & \best{77.6}
    & 89.9
\\ 
vegetation    & 78.4
    & 79.4
    & \best{79.7}
    & 79.5
    & 81.0
    & \best{81.7}
    & 86.5
\\ 

\midrule

mean   & 46.4
    & 47.3
    & \best{47.7}    
    & 48.0
    & 48.9
    & \best{49.3}
    & 78.7
\\ 

\midrule

\multicolumn{8}{l}{\textbf{Superclasses}}  \\
\midrule

object   & 5.3
    & 6.0
    & \best{6.2}
    & 6.6
    & 6.4
    & \best{7.6}
\\ 
vehicle   & 79.9 
    & 81.8
    & \best{83.3}
    & 80.5
    & 82.4
    & \best{83.0}
\\ 
human   & 67.1
    &  67.9
    & \best{68.2}
    & 70.8
    &  72.1
    & \best{72.7}
\\ 
ground   & 90.1
    & 90.6
    & \best{90.8}
    & 90.5
    & 90.9
    & \best{91.0}
\\ 
nature   & 76.7
    & 77.6
    & \best{77.9}
    & 77.3
    & 78.2 
    & \best{78.5}
\\ 
structure   & 74.4
    & 75.8
    & \best{76.3}
    & 74.6
    & 76.5
    & \best{77.6}
\\
\midrule
mean   & 65.6
    & 66.1
    & \best{67.1}
    & 66.7
    & 67.8
    & \best{68.4}
\\

\bottomrule
\end{tabular}
\end{center}
\fboxsep 1.5pt
\caption{\textbf{Classwise semantic segmentation results (IoU\%) of \ours on \ns \underline{val set}} when finetuning a 3D network \cite{waffleiron, scalr} on labelings $\Ltim$ or $\Lcomb$, and iterating training interleaved with time-based consolidation. 
``Full supp.'' is when training with ground-truth labels \cite{waffleiron, scalr}. Best results are in \smash{\colorbox{blue!17}{blue}}.}
\label{table:class_wise_iou_ns}
\vspace*{-0cm}
\end{table*}%
These tables detail the results reported in Table~3 of the main paper. They also provide classwise results for super-classes (cf.\,\cref{sec:sc-results}).

\begin{table*}[t!]
\begin{center}
\def\best{\cellcolor{blue!17}}
\begin{tabular}{l|rrr|rrr|r}
\toprule
\textbf{\sk}
    & \multicolumn{3}{c|}{$\Ltim$}
    & \multicolumn{3}{c|}{$\Lcomb$}
    & Full\hspace{0.5ex}
\\
 \cmidrule(lr){2-4}  \cmidrule(lr){5-7}
\textbf{Classes} (default) & 1\textsuperscript{st} iter.\hspace{-1ex}
    & 2\textsuperscript{nd} iter.\hspace{-1ex}
    & 3\textsuperscript{rd} iter.\hspace{-1ex}
    & 1\textsuperscript{st} iter.\hspace{-1ex}
    & 2\textsuperscript{nd} iter.\hspace{-1ex}
    & 3\textsuperscript{rd} iter.\hspace{-1ex}
    & supp.
\\
\midrule
car
    & 77.1
    & \best 79.9
    & 79.6
    & 79.1
    & \best 80.8
    & \best 80.8
    & 94.7
\\ 
bicycle    & 35.6
    & 37.4
    & \best 37.5
    & 33.3
    & \best 35.5
    & 34.7
    & 44.7
\\ 
motorcycle    & 58.4
    & 61.3
    & \best 63.4
    & \best 58.0
    & 56.0
    & 56.7
    & 64.8
\\ 
truck    & 10.5
    & \best 10.6
    & 10.3
    & \best 17.3
    & 16.5
    & 16.0
    & 82.0
\\ 
other-vehicle    & \best 4.8
    & 2.7
    & 2.3
    & 30.8
    & 32.3
    & \best 34.1
    & 37.4
\\ 
person    & 52.8
    & 54.5
    & \best 55.0
    & 55.0
    & \best 56.7
    & 56.5
    & 71.4
\\ 
bicyclist    & 0.0
    & 0.0
    & 0.0
    & 0.0
    & 0.0
    & 0.0
    & 85.9
\\ 
motorcyclist    & 0.0
    & 0.0
    & 0.0
    & 0.0
    & 0.0
    & 0.0
    & 0.0
\\ 
road    & 75.6 
    & \best 76.0
    & 75.5
    & 75.5
    & \best 75.7
    & 75.3
    & 95.7
\\ 
parking    & 0.0
    & 0.0
    & 0.0
    & 0.0
    & 0.0
    & 0.0
    & 49.2
\\ 
sidewalk    & 55.2
    & \best 55.9
    & 55.3
    & 55.2
    & \best 55.3
    & 54.9
    & 83.6
\\ 
other-ground    & 0.0
    & 0.0
    & 0.0
    & 0.0
    & 0.0
    & 0.0
    & 0.2
\\ 
building    & 69.2
    & 73.7
    & \best 76.0
    & 69.3
    & 73.9
    & \best 76.5
    & 89.3
\\ 
fence    & 8.0
    & 8.9
    & \best 9.3 
    & 7.3
    & \best 9.2
    & 9.0
    & 56.0
\\ 
vegetation    & 80.6
    & 83.1
    & \best 84.1
    & 80.5
    & 82.8
    & \best 83.8
    & 88.2
\\ 
trunk    & 0.0
    & 0.0
    & 0.0
    & 0.0
    & 0.0
    & 0.0
    & 69.9
\\ 
terrain    & 66.7
    & \best 67.9
    & 67.8
    & 67.6
    & \best 67.9
    & 66.9
    & 74.6
\\ 
pole    & 0.0
    & 0.0
    & 0.0
    & 0.0
    & 0.0
    & 0.0
    & 66.0
\\ 
traffic-sign    & \best 16.2
    & 15.4
    & 15.5
    & 21.4
    & \best 22.9
    & \best 22.9
    & 51.5
\\ 
\midrule

mean   & 32.1 
    & 33.0
    & \best 33.2
    & 34.2
    & 35.0
    & \best 35.2
    & 63.4
\\ 

\midrule

\multicolumn{8}{l}{\textbf{Superclasses}} \\
\midrule

object   & 7.6
    & 8.4
    & \best 8.8
    & 7.2
    & \best 9.1
    & 8.9
\\ 
vehicle   & 83.9 
    & 86.5
    & \best 86.6
    & 84.0
    & 86.6
    & \best 87.0
\\ 
human   & 65.6 
    &  \best 66.0
    & 65.7
    & 63.8
    & \best 63.9
    & 63.1
\\ 
ground   & 84.0
    & \best 84.3 
    & 84.0
    & \best 84.3
    & 84.1
    & 83.6
\\ 
nature   & 84.4 
    & 85.9
    & \best 86.3
    & 84.3
    & 85.6 
    & \best 85.7
\\ 
structure   & 69.2 
    & 73.7
    & \best 76.0
    & 69.3
    & 73.9
    & \best 76.5
\\
\midrule
mean   & 65.8 
    & 67.5
    & \best 67.9
    & 65.6
    & 67.2
    & \best 67.4
\\

\bottomrule
\end{tabular}
\end{center}
\fboxsep 1.5pt
\caption{\textbf{Classwise semantic segmentation results (IoU\%) of \ours on \sk \underline{val set}} when finetuning a 3D network \cite{waffleiron, scalr} on labelings $\Ltim$ or $\Lcomb$, and iterating training interleaved with time-based consolidation. 
``Full supp.'' is when training with ground-truth labels \cite{waffleiron, scalr}. Best results (besides full supervision) for each labeling are with a \smash{\colorbox{blue!17}{blue}} background.}
\label{table:class_wise_iou_sk}
\end{table*}

For \ns, while the IoU\% of most classes keeps on increasing as we iterate more, only a few classes see their performance stagnate or slightly drop. Therefore, the overall performance (mIoU\%) is clearly increasing even after 3 iterations.

For \sk, where the overall performance is lower and increasing at a lower rate as we iterate, compared to \ns, a higher number of classes have a best performance at the 1\textsuperscript{st} or 2\textsuperscript{nd} iteration. Still, the overall performance after the 3\textsuperscript{rd} iteration remains the best one.

The results for the method we call \ours (Table~5 of the main paper) correspond to three iterations.

\section{Quantity and quality of labeled points for training}
\label{sec:labeled_points}

Table \ref{table:trainpts_ns} below lists the number of labeled points per class in the \ns training set for the different labelings reported in Table~4 of the main paper. Colors in these tables illustrate the choices made in Eq.\,(5) to select labels that are more likely to be trustworthy, among labels from $\Ltim$ or $\Laug$.

\begin{table*}[h!]
\def\fovgt{Fo\!V\!\textsubscript{gt}}
\begin{center}
\setlength{\tabcolsep}{4pt}
\def\gray{\cellcolor{gray!17}}
\def\sel{\cellcolor{green!17}}
\def\low{\cellcolor{yellow!30}}
\resizebox{1\textwidth}{!}{%
\begin{tabular}{l|@{~}rrrrrr|r|rrr}
\toprule
Class				& $\Lini$& $\Ltim$	& $\Labc$& $\Laug$	& \gray \hspace*{-1ex}$\Naug/\Ntim$\hspace*{-1ex} & $\Lcomb$ & \fovgt &	\hspace{1ex}1\textsuperscript{st}\,iter.\hspace{-1ex} & 2\textsuperscript{nd}\,iter.\hspace{-1ex} & 3\textsuperscript{rd}\,iter.\hspace{-1ex}\\
\midrule
barrier				& 2.0	& \sel 2.1	& 0.5	& 0.4		& \it \low 	0.20	& 2.1	& 7.7	& 4.8	& 4.8	& 4.7\\
bicycle				& 0.3	& \sel 0.3	& 0.1	& \low 0.1	& \it \gray 0.46	& 0.3	& 0.1	& 0.5	& 0.6	& 0.6 \\
bus					& 3.7	& 3.7		& 2.5	& \sel 2.4	& \it \gray 0.65	& 2.4	& 3.8	& 8.1	& 7.7	& 7.9 \\
car					& 27.4	& 27.5		& 24.4	& \sel 24.0	& \it \gray 0.87	& 24.0	& 32.0	& 93.8	& 102.2	& 103.7\\
construction vehicle& 0.1	& \sel 0.1	& 0.0	& \low 0.0	& \it \low 	0.10	& 0.1	& 1.3	& 0.4	& 0.3	& 0.3 \\
motorcycle			& 0.3	& \sel 0.3	& 0.2	& \low 0.2	& \it \gray 0.70	& 0.3	& 0.3	& 0.3	& 0.3	& 0.3 \\
pedestrian			& 2.3	& 2.2		& 1.8	& \sel 1.7	& \it \gray 0.75	& 1.7	& 1.9	& 1.8	& 1.7	& 1.7 \\
traffic cone		& 0.2	& \sel 0.2	& 0.1	& \low 0.1	& \it \gray 0.45	& 0.2	& 0.6	& 0.5	& 0.5	& 0.5\\
trailer				& 0.2	& \sel 0.1	& 0.0	& \low 0.0	& \it \low 	0.06	& 0.1	& 4.5	& 0.9	& 1.1	& 1.2 \\
truck				& 20.4	& 20.4		& 15.7	& \sel 15.5	& \it \gray 0.76	& 15.5	& 13.3	& 58.7	& 54.8	& 49.6 \\
driveable surface	& 156.3	& 149.4		& 142.1	& \sel 134.3& \it \gray 0.90	& 134.3	& 260.9	& 298.0	& 299.6	& 298.9 \\
other flat			& 0.1	& \sel 0.1	& 0.1	& \low 0.1	& \it \gray 0.47	& 0.1	& 6.6	& 0.7	& 0.3	& 0.3 \\
sidewalk			& 51.7	& 50.0		& 38.4	& \sel 36.5	& \it \gray 0.73	& 36.5	& 57.6	& 69.5	& 68.2	& 68.1 \\
terrain				& 38.6	& 37.7		& 28.1	& \sel 27.0	& \it \gray 0.72	& 27.0	& 56.7	& 54.1	& 50.1	& 47.5 \\
manmade				& 128.6	& 128.5		& 109.9	& \sel 108.7& \it \gray 0.85	& 108.7	& 146.5	& 168.9	& 152.2	& 147.0 \\
vegetation			& 94.8	& 92.4		& 79.0	& \sel 77.1	& \it \gray 0.83	& 77.1	& 100.6	& 215.8	& 232.3	& 244.4 \\
\bottomrule
\end{tabular}%
}
\end{center}
\fboxsep 1.5pt
\caption{\textbf{Quantity of labeled points used in \ns \underline{train set}.} The number of points is given in millions. Values under 50k appear as ``0.0''. Values not meeting the conditions in Eq.\,(4) are indicated with a \colorbox{yellow!30}{yellow} background. Labels from $\Ltim$ or $\Laug$ that are selected in $\Lcomb$ as specified in Eq.\,(5) are shown with a \colorbox{green!17}{green} background. Column ``\fovgt'' gives the number of points in the field of view of the cameras for each class according to the ground truth, for comparison purposes. The following columns (1\textsuperscript{st} to 3\textsuperscript{rd}\,iter.) provide results of trained 3D networks and thus concern all scanned points, not just those visible from the cameras.}
\label{table:trainpts_ns}
\vspace*{-3pt}
\end{table*}

Table \ref{table:quality_ns} provides the classwise quality (IoU\%) of labeled points in the \ns training set for the different labelings reported in Table~4 of the main paper. As in Table \ref{table:trainpts_ns}, color in this table illustrates the choices made in Eq.\,(5) to select labels from $\Ltim$ or $\Laug$ that are more likely to be trustworthy. Note that points that receive no label with our methods are excluded when computing the metric. Hence, from one column to another (different labelings), the set of points on which the IoU is computed differs.

\begin{table*}[h!]
\begin{center}
\resizebox{0.99\textwidth}{!}{
\def\sel{\cellcolor{green!17}}
\def\bst{\cellcolor{blue!30}}
\begin{tabular}{l|rrrrr|rrr|r@{\hspace*{3ex}}}
\toprule
Class				& $\Lini$& $\Ltim$	& $\Labc$& $\Laug$	& \hspace*{-2ex}$\Lcomb$ & \hspace{-1ex}
1\textsuperscript{st}\,iter.\hspace{-1ex} & 2\textsuperscript{nd}\,iter.\hspace{-1ex} & 3\textsuperscript{rd}\,iter.\hspace{0ex} &
F.\,supp.\hspace{-2ex} \\
\midrule
barrier				& 4.2	& \sel	4.6		& 2.2	& 		2.1		& 5.4	& 5.0	& 4.8	& 4.8	& 79.2 \\
bicycle				& 21.0	& \sel	22.0	& 26.5	& 		28.0	& 23.6	& 17.8	& 15.7	& 15.9	& 53.2 \\
bus					& 57.5	& 		57.5	& 74.4	& \sel	74.9	& 74.9	& 66.4	& 67.2	& 67.7	& 92.5 \\
car					& 63.4	& 		63.5	& 69.8	& \sel	70.5	& 70.4	& 73.3	& 76.1	& 77.3	& 88.1 \\
construction vehicle& 5.4	& \sel	5.2		& 1.4	& 		1.1		& 7.2	& 11.9	& 12.1	& 12.4	& 50.4 \\
motorcycle			& 41.8	& \sel	43.0	& 50.5	& 		52.2	& 48.5	& 57.2	& 59.1	& 59.6	& 87.8 \\
pedestrian			& 44.0	& 		47.0	& 52.5	& \sel	56.6	& 56.4	& 70.5	& 71.7	& 72.0	& 83.4 \\
traffic cone		& 24.3	& \sel	26.0	& 25.8	& 		26.8	& 34.6	& 45.5	& 48.0	& 48.8	& 70.3 \\
trailer				& 0.0	& \sel	0.0		& 0.0	& 		0.0		& 0.0	& 0.0	& 0.0	& 0.0	& 73.5 \\
truck				& 41.3	& 		41.5	& 48.4	& \sel	49.2	& 49.1	& 47.7	& 48.4	& 48.7	& 84.1 \\
driveable surface	& 84.4	& 		84.2	& 90.7	& \sel	91.0	& 91.0	& 89.8	& 89.8	& 89.7	& 96.9 \\
other flat			& 0.0	& \sel	0.0		& 0.0	& 		0.0		& 0.0	& 0.0	& 0.0	& 0.0	& 73.3 \\
sidewalk			& 48.4	& 		48.6	& 57.9	& \sel	59.0	& 59.0	& 53.5	& 54.3	& 54.8	& 75.5 \\
terrain				& 52.3	& 		52.6	& 65.2	& \sel	66.6	& 66.4	& 57.0	& 57.5	& 57.5	& 75.4 \\
manmade				& 69.5	& 		70.3	& 77.8	& \sel	79.1	& 78.3	& 74.2	& 75.7	& 76.5	& 89.9 \\
vegetation			& 71.4	& 		72.1	& 79.8	& \sel	81.0	& 81.0	& 79.3	& 80.6	& 81.2	& 86.5 \\
\midrule
mean				& 39.3	& 	39.9	& 45.1	& 46.1	& 46.6	& 46.8	& 47.6	& 47.9	& 78.7 \\
\bottomrule
\end{tabular}
}
\end{center}
\fboxsep 1.5pt
\caption{\textbf{Quality (IoU\%) of labeled points used in \ns \underline{train set}.} 
Labels from $\Ltim$ or \smash{$\Laug$} selected in $\Lcomb$ are shown with a \colorbox{green!17}{green} background. Columns ``1\textsuperscript{st}\,iter.'' to ``3\textsuperscript{rd}\,iter.'' provide results of trained 3D networks and thus concern all scanned points, not just those visible from the cameras as in the previous columns. ``F.\,supp.'' is when training with ground-truth labels \cite{waffleiron, scalr}.}
\label{table:quality_ns}
\vspace*{-0cm}
\end{table*}

\section{Text prompts}
\label{sec:prompts}

In this section, we provide detailed class-specific prompts for both our minimal and rich text prompt settings. Table~\ref{tab:text_prompts} lists the \emph{minimal} prompts employed in our final models for \ns and \sk. Additionally, Table~\ref{tab:text_prompts_rich_ns} presents the class-specific rich prompts used for \ns in the benchmarking of 2D open-vocabulary VLMs, as reported in Table 1 of the main paper.

\begin{table*}[!h]
    \centering
    \scriptsize
    \caption{Our classwise \textit{minimal} text prompts used to obtain segmentation maps from OpenSeed~\cite{openseed}.
    }
    \setlength\extrarowheight{4pt}
    \begin{tabular}{p{0.2\textwidth}|p{0.73\textwidth}}
    \toprule
    Class & Text prompts \\
    \midrule

    \multicolumn{2}{c}{\ns} \\
    \midrule
    \texttt{pedestrian} & \texttt{person}, \texttt{pedestrian} \\
    
    \texttt{bicycle} & \texttt{bicycle} \\
    \texttt{bus} & \texttt{bus} \\
    \texttt{car} & \texttt{car} \\
    \texttt{construction vehicle} & \texttt{bulldozer}, \texttt{excavator}, \texttt{concrete mixer}
    , \texttt{crane},  \texttt{dump truck}   \\											
    \texttt{motorcycle} & \texttt{motorcycle} \\
    
    \texttt{trailer} & \texttt{trailer},  \texttt{sem trailer},  \texttt{cargo container},  \texttt{shipping container},  \texttt{freight container}  \\
    \texttt{truck} & \texttt{truck} \\
    \texttt{barrier} & \texttt{barrier}, \texttt{barricade} \\
    \texttt{traffic cone} & \texttt{traffic cone} \\
    \texttt{driveable surface} & \texttt{road}\\ 																							
    \texttt{other flat} & \texttt{curb}, \texttt{traffic island}, \texttt{traffic median}  \\
    \texttt{sidewalk} & \texttt{sidewalk}\\
    
    \texttt{terrain} & \texttt{terrain}, \texttt{grass}, \texttt{grassland},  \texttt{lawn},   \texttt{meadow},  \texttt{turf}, \texttt{sod} \\

    \texttt{manmade} & \texttt{building}, \texttt{wall}, \texttt{pole}, \texttt{awning} \\
    \texttt{vegetation} & \texttt{tree}, \texttt{trunk}, \texttt{tree trunk}, \texttt{bush}, \texttt{shrub}, \texttt{plant}, \texttt{flower}, \texttt{woods} \\
\midrule
    
    \multicolumn{2}{c}{\sk} \\
    \midrule
    \texttt{car} & \texttt{car}\\
    \texttt{bicycle} & \texttt{bicycle} \\
    \texttt{motorcycle} & \texttt{motorcycle} \\
    \texttt{truck} &\texttt{truck} \\
    \texttt{other vehicle} & \texttt{trailer}, \texttt{semi trailer}, \texttt{cargo container}, \texttt{shipping container}, \texttt{freight container}, \texttt{caravan},  \texttt{bus}, \texttt{bulldozer}, \texttt{excavator}, \texttt{concrete mixer}, \texttt{crane}, \texttt{dump truck},
    \texttt{train},  \texttt{tram}\\
    \texttt{person} & \texttt{person}, \texttt{pedestrian} \\
    \texttt{bicyclist} & \texttt{bicyclist}, \texttt{cyclist} \\
    \texttt{motorcyclist} & \texttt{motorcyclist}\\
    \texttt{road} & \texttt{road}\\
    \texttt{parking} & \texttt{parking}, \texttt{parking lot} \\
    \texttt{sidewalk} & \texttt{sidewalk}, \texttt{curb}, \texttt{bike path}, \texttt{walkway}, 
    \texttt{pavement},\texttt{footpath},\texttt{footway},\texttt{boardwalk}, \texttt{driveway}\\														
    \texttt{other ground} & \texttt{water}, \texttt{river}, \texttt{lake}, \texttt{watercourse}, \texttt{waterway}, \texttt{canal}, \texttt{ditch}, \texttt{rail track}, \texttt{traffic island}, \texttt{traffic median}, \texttt{median strip}, \texttt{roadway median}, \texttt{central reservation} \\																		
    \texttt{building} & \texttt{building}, \texttt{house}, \texttt{garage}, \texttt{wall}, \texttt{railing}, \texttt{stairs}, 
    \texttt{awning}, \texttt{roof}, \texttt{bridge}\\
    \texttt{fence} & \texttt{fence}, \texttt{barrier}, \texttt{barricade} \\    
    \texttt{vegetation} & \texttt{tree}, \texttt{bush}, \texttt{shrub}, \texttt{plant}, \texttt{flower} \\
    \texttt{trunk} &  \texttt{tree trunk}, \texttt{trunk}, \texttt{woods} \\																								
    \texttt{terrain} & \texttt{terrain}, \texttt{grass}, \texttt{soil}, \texttt{grassland}, \texttt{hill}, 
    \texttt{sand}, \texttt{gravel}, \texttt{lawn}, \texttt{meadow}, \texttt{garden}, \texttt{earth}, \texttt{peeble}, \texttt{rock} \\
    \texttt{pole} & \texttt{pole} \\
    \texttt{traffic sign} & \texttt{traffic-sign} \raisebox{-5pt}{}\\

    \bottomrule
    \end{tabular}
    \label{tab:text_prompts} 
\end{table*}

\begin{table*}[h]
    \centering
    \scriptsize
    \caption{Our \textit{rich} text prompts used to obtain segmentation maps from OpenSeed~\cite{openseed}
    on \ns.}
    \setlength\extrarowheight{4pt}
    \begin{tabular}{p{0.2\textwidth}|p{0.74\textwidth}}
    \toprule
    Class & Text prompts \\
    \midrule

    \multicolumn{2}{c}{\ns} \\
    \midrule

    \texttt{pedestrian} & \texttt{pedestrian}, \texttt{person}, \texttt{human}, \texttt{man}, \texttt{woman}, \texttt{adult}, \texttt{child}, \texttt{stroller}, \texttt{wheelchair} \\
    
    \texttt{bicycle} & \texttt{bicycle} \\
    
    \texttt{bus} & \texttt{bus}, \texttt{autobus}, \texttt{motorbus}, \texttt{omnibus}, \texttt{double-decker}, \texttt{jitney}, \texttt{minibus}, \texttt{motor coach}, \texttt{school bus}, \texttt{tour bus}, \texttt{sightseeing bus}, \texttt{shuttle bus}, \texttt{shuttle}, \texttt{bendy bus} \\
    
    \texttt{car} & \texttt{car}, \texttt{automobile}, \texttt{van}, \texttt{SUV}, \texttt{sedan}, \texttt{hatchback}, \texttt{wagon}, \texttt{minivan}, \texttt{convertible}, \texttt{jeep},\texttt{ambulance} \\
    
    \texttt{construction vehicle} & \texttt{bulldozer}, \texttt{excavator}, \texttt{concrete mixer}, \texttt{crane}, \texttt{dump truck}, \texttt{dozer}, \texttt{digger}, \texttt{dumper}, \texttt{tipper truck}, \texttt{tipper lorry}, \texttt{front loader}, \texttt{loader}, \texttt{backhoe}, \texttt{trencher}, \texttt{compactor}, \texttt{forklift} \\
    
    \texttt{motorcycle} & \texttt{motorcycle}, \texttt{motorbike}, \texttt{choper}, \texttt{motor scooter}, \texttt{scooter}, \texttt{vespa} \\
    
    \texttt{trailer} & \texttt{trailer}, \texttt{semi trailer}, \texttt{cargo container}, \texttt{shipping container}, \texttt{freight container}  \\
    
    \texttt{truck} & \texttt{truck}, \texttt{pickup}, \texttt{pickup truck}, \texttt{lorry}, \texttt{autotruck}, \texttt{semi}, \texttt{semi-tractor}, \texttt{semi-truck}, \texttt{motortruck} \\
    
    \texttt{barrier} & \texttt{barrier}, \texttt{jersey barrier}, \texttt{barricade} \\
    
    \texttt{traffic cone} & \texttt{traffic cone}, \texttt{safety pylon}, \texttt{road cone}, \texttt{highway cone}, \texttt{safety cone}, \texttt{caution cone}, \texttt{channelizing device}, \texttt{construction cone}, \texttt{cone}, \texttt{road delineator}  \\
    
    \texttt{driveable surface} & \texttt{road}, \texttt{parking}, \texttt{parking lot}, \texttt{highway}, \texttt{expressway}, \texttt{freeway}, \texttt{thoroughfare}, \texttt{route}, \texttt{thruway}, \texttt{turnpike}, \texttt{roadway}, \texttt{macadam}, \texttt{lane} \\ 																							
    \texttt{other flat} & \texttt{water}, \texttt{river}, \texttt{lake}, \texttt{watercourse}, \texttt{waterway}, \texttt{canal}, \texttt{ditch}, \texttt{railway track}, \texttt{traffic island}, \texttt{traffic median}, \texttt{median strip}, \texttt{roadway median}, \texttt{central reservation}  \\
    
    \texttt{sidewalk} & \texttt{sidewalk}, \texttt{curb}, \texttt{bike path}, \texttt{walkway}, \texttt{pavement}, \texttt{footpath}, \texttt{footway}, \texttt{boardwalk} \\
    
    \texttt{terrain} & \texttt{terrain}, \texttt{grass}, \texttt{grassland}, \texttt{hill}, \texttt{soil}, \texttt{sand}, \texttt{gravel}, \texttt{lawn}, \texttt{meadow}, \texttt{garden}, \texttt{earth}, \texttt{peeble}, \texttt{rock} \\

    \texttt{manmade} & \texttt{building}, \texttt{skyscraper}, \texttt{house}, \texttt{wall}, \texttt{stairs}, \texttt{awning}, \texttt{roof}, \texttt{pole}, \texttt{streetlight}, \texttt{traffic light}, \texttt{traffic sign}, \texttt{bench}, \texttt{fire hydrant}, \texttt{hydrant}, \texttt{guard rail}, \texttt{guardrails}, \texttt{bollard}, \texttt{fence}, \texttt{railing}, \texttt{drainage}, \texttt{flag}, \texttt{banner}, \texttt{street sign}, \texttt{circuit box}, \texttt{traffic signal}, \texttt{stoplight}, \texttt{lamppost}, \texttt{bus shelter}, \texttt{bus stop}, \texttt{bin}, \texttt{mailbox}, \texttt{newspaper box}, \texttt{newsbox} \\
    
    \texttt{vegetation} & \texttt{tree}, \texttt{bush}, \texttt{plant}, \texttt{shrub}, \texttt{potted plant}, \texttt{hedge}, \texttt{branch}, \texttt{tree trunk}, \texttt{flower}, \texttt{woods}, \texttt{forest} \raisebox{-5pt}{}\\

    \bottomrule
    \end{tabular}
    \label{tab:text_prompts_rich_ns} 
\end{table*}

\section{Augmentations}
\label{sec:augmentation}

The following Table~\ref{tab:augmentation_ns_selected} presents detailed classwise results of zero-shot OpenSeeD~\cite{openseed} applied to the \ns dataset, before and after using our set of 10 distinct data augmentation methods from the Albumentations library \cite{albumentations}. We observe that the results remain largely consistent with the original, which helps us eliminate obviously incorrect labels.
We also present the results of several other data augmentations in Table~\ref{tab:augmentation_ns_not_selected}. These augmentations perform significantly worse than the no-augmentation baseline and result in a substantial loss of points.

\begin{table*}[h!]
\def\fovgt{Fo\!V\!\textsubscript{gt}}
\begin{center}
\setlength{\tabcolsep}{4pt}
\def\gray{\cellcolor{gray!17}}
\def\sel{\cellcolor{green!17}}
\def\low{\cellcolor{yellow!30}}
\resizebox{1\textwidth}{!}{%
\begin{tabular}{l|c|cccccccccc}
\toprule
\multirow{2}{*}{Class} &	no  & {horizontal} & hue & \multirow{2}{*}{blur} & color & auto- & \multirow{2}{*}{sharpen} & chromatic  & \multirow{2}{*}{emboss} &  fancy  & \multirow{2}{*}{clahe}  \\
& aug. & flip & saturation &   & jitter  & contrast &  &  aberration&  & pca &  \\
\midrule
barrier				& 1.6  & 1.9  & 2.1  & 1.7   & 1.7  & 2.6  & 1.7  & 1.6  & 1.7  & 1.6  & 1.4 \\
bicycle				& 24.3 & 24.2 & 23.9 & 23.5  & 24.3 & 24.0 & 24.2 & 23.8 & 23.7 & 24.3 & 24.0 \\
bus					& 56.7 & 57.2 & 54.6 & 57.4  & 56.2 & 49.7 & 53.0 & 56.0 & 56.6 & 56.8 & 55.6 \\
car					& 47.9 & 48.0 & 48.0 & 48.4  & 48.1 & 46.3 & 47.8 & 47.7 & 47.2 & 47.9 & 48.3 \\
construction vehicle&  5.2 & 5.1  & 4.0  & 3.7   & 4.2  & 3.1  & 3.7  & 3.3  & 3.8  & 5.2  & 4.2 \\
motorcycle			& 31.2 & 31.2 & 30.7 & 31.2  & 30.8 & 29.9 & 30.8 & 30.6 & 31.1 & 31.2 & 29.9 \\
pedestrian			& 40.4 & 40.5 & 40.3 & 40.6  & 40.6 & 39.5 & 40.2 & 40.3 & 39.7 & 40.4 & 39.0 \\
traffic cone		& 20.5 & 20.4 & 19.2 & 21.3  & 20.3 & 22.8 & 18.3 & 19.8 & 19.0 & 20.5 & 16.6 \\
trailer				& 0.1  & 0.1  & 0.2  & 0.1   & 0.1  & 0.2  & 0.2  & 0.1  & 0.1  & 0.2  & 0.1 \\
truck				& 45.7 & 45.5 & 45.0 & 46.1  & 45.8 & 44.3 & 45.2 & 45.1 & 45.1 & 45.7 & 46.0 \\
driveable surface	& 50.6 & 50.6 & 50.3 & 50.2  & 50.4 & 50.2 & 50.2 & 50.1 & 50.3 & 50.6 & 50.4 \\
other flat			& 0.0  & 0.0  & 0.0  & 0.0   & 0.0  & 0.0  & 0.0  & 0.0  & 0.0  & 0.0  & 0.0 \\
sidewalk			& 34.4 & 34.4 & 33.9 & 33.6  & 34.1 & 33.8 & 34.2 & 33.0 & 33.8 & 34.4 & 33.9 \\
terrain				& 43.9 & 43.8 & 42.7 & 43.4  & 43.7 & 44.0 & 43.6 & 43.3 & 43.2 & 43.8 & 43.6 \\
manmade				& 66.7 & 66.7 & 65.9 & 66.9  & 66.4 & 65.2 & 65.7 & 66.4 & 65.7 & 66.7 & 66.6 \\
vegetation			& 66.7 & 66.6 & 65.7 & 67.4  & 66.2 & 65.6 & 64.4 & 66.8 & 64.9 & 66.7 & 67.7 \\
\midrule
mean IoU			& 33.4 & 33.5 & 32.9 & 33.4  & 33.3 & 32.5 & 32.6 & 32.9 & 32.8 & 33.4 & 32.9 \\
mean Acc			& 43.4 & 43.4 & 42.8 & 43.3  & 43.1 & 42.8 & 42.4 & 42.7 & 42.4 & 43.4 & 42.9 \\

\bottomrule
\end{tabular}%
}
\end{center}
\fboxsep 1.5pt
\caption{\textbf{Classwise results on the \ns validation set, before and after applying our 10 selected data augmentations.} These augmentations yield results comparable to the no-augmentation baseline and help eliminate unreliable predictions.}
\label{tab:augmentation_ns_selected}
\vspace*{-3pt}
%
\def\fovgt{Fo\!V\!\textsubscript{gt}}
\begin{center}
\setlength{\tabcolsep}{4pt}
\def\gray{\cellcolor{gray!17}}
\def\sel{\cellcolor{green!17}}
\def\low{\cellcolor{yellow!30}}
\resizebox{1\textwidth}{!}{%
\begin{tabular}{l|c|ccccccccc}
\toprule

\multirow{2}{*}{Class} & no  & salt-and & gauss & random & random & random & glass & shot  & \multirow{2}{*}{superpixels} &  vertical \\
& aug. &  pepper & noise & sun-flare  & snow  & rain & blur &  noise &  & flip  \\
\midrule
barrier				& 1.6 & 1.6   & 1.0  & 1.3   & 1.3  & 1.1  & 0.5  & 0.9  & 0.9  & 0.0  \\
bicycle				& 24.3 & 21.6 & 17.8 & 24.0  & 20.8 & 23.5 &14.3  & 14.4 & 14.1 & 14.8  \\
bus					& 56.7 & 46.9 & 50.0 & 45.5  & 50.5 & 42.9 & 35.2 & 37.3 & 36.8 & 22.8  \\
car					& 47.9 & 48.2 & 48.2 & 45.3  & 48.0 & 46.2 & 47.5 & 45.5 & 39.6 & 36.5  \\
construction vehicle& 5.2 & 1.2   & 1.0  & 2.2   & 0.9  & 2.4  & 0.0  & 1.0  & 0.7  & 0.0  \\
motorcycle			& 31.2 & 27.9 & 27.2 & 29.2  & 27.6 & 26.6 & 18.5 & 23.1 & 21.0 & 5.4  \\
pedestrian			& 40.4 & 37.4 & 36.3 & 39.7  & 39.7 & 40.1 & 37.7 & 31.2 & 33.4 & 31.6  \\
traffic cone		& 20.5 & 17.8 & 19.0 & 18.2  & 14.7 & 20.5 & 17.6 & 17.2 & 14.0 & 0.0  \\
trailer				& 0.1 & 0.0   & 0.0  & 0.0   & 0.1  & 0.0  & 0.0  & 0.0  & 0.0  & 0.7  \\
truck				& 45.7 & 43.0 & 45.5 & 42.5  & 44.0 & 39.9 & 37.8 & 35.4 & 35.4 & 22.3  \\
driveable surface	& 50.6 & 48.4 & 48.7 & 48.9  & 48.7 & 48.2 & 45.1 & 46.6 & 46.6 & 17.1  \\
other flat			& 0.0 & 0.0   & 0.0  & 0.0   & 0.0  & 0.0  & 0.0  & 0.0  & 0.0  & 0.0  \\
sidewalk			& 34.4 & 29.8 & 28.9 & 31.0  & 28.8 & 27.4 & 16.5 & 25.2 & 26.4 & 7.5  \\
terrain				& 43.9 & 41.0 & 41.6 & 38.9  & 37.6 & 42.6 & 40.0 & 37.3 & 40.5 & 36.6 \\
manmade				& 66.7 & 63.2 & 61.3 & 62.7  & 65.3 & 60.1 & 59.9 & 56.3 & 61.4 & 40.3  \\
vegetation			& 66.7 & 60.8 & 60.8 & 59.6  & 64.7 & 60.2 & 62.8 & 55.2 & 59.0 & 60.0  \\
\midrule
mean IoU			& 33.4 & 30.5 & 30.4 & 30.5  & 30.7 & 30.1 & 27.0 & 26.6 & 26.8 & 18.4  \\
mean Acc			& 43.4 & 39.3 & 39.8 & 40.5  & 40.1 & 40.0 & 35.0 & 35.4 & 34.6 & 27.1  \\

\bottomrule
\end{tabular}%
}
\end{center}
\fboxsep 1.5pt
\caption{\textbf{Classwise results on the \ns validation set, comparing performance before and after applying several additional data augmentations.} These augmentations significantly degraded performance compared to the baseline without any augmentation. Due to their negative impact and deviation from the original scores, we excluded them from our pipeline to ensure sufficient points in the scenes.}
\label{tab:augmentation_ns_not_selected}
\vspace*{-3pt}
\end{table*}

\begin{figure*}[h]
\centering
\begin{minipage}{0.75\linewidth}
\centering Color code used for nuScenes data
\end{minipage}
\\
\fbox{\includegraphics[width=.75\linewidth,trim={0cm 0cm 0cm 3cm},clip]{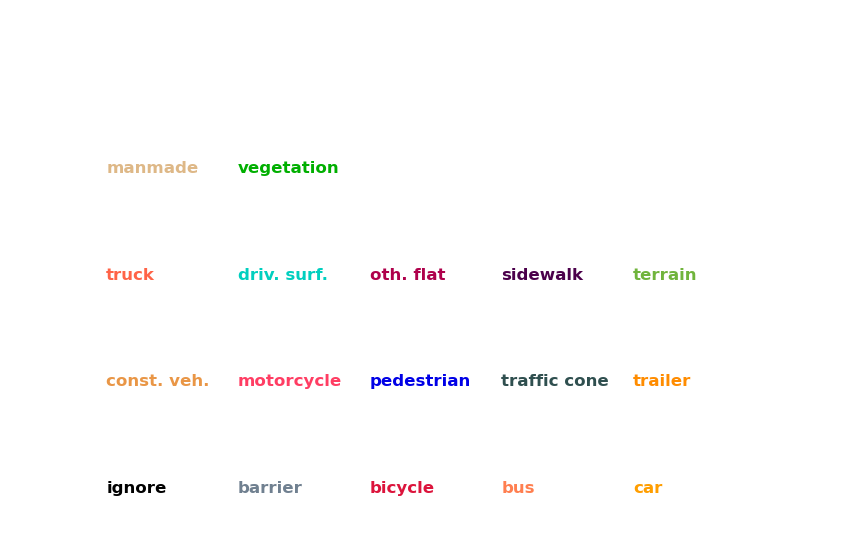}}
\vspace{5mm}
\\
\begin{minipage}{0.75\linewidth}
\centering Color code used for SemanticKITTI data
\end{minipage}
\\
\fbox{\includegraphics[width=.75\linewidth,trim={0cm 0cm 0cm 3cm},clip]{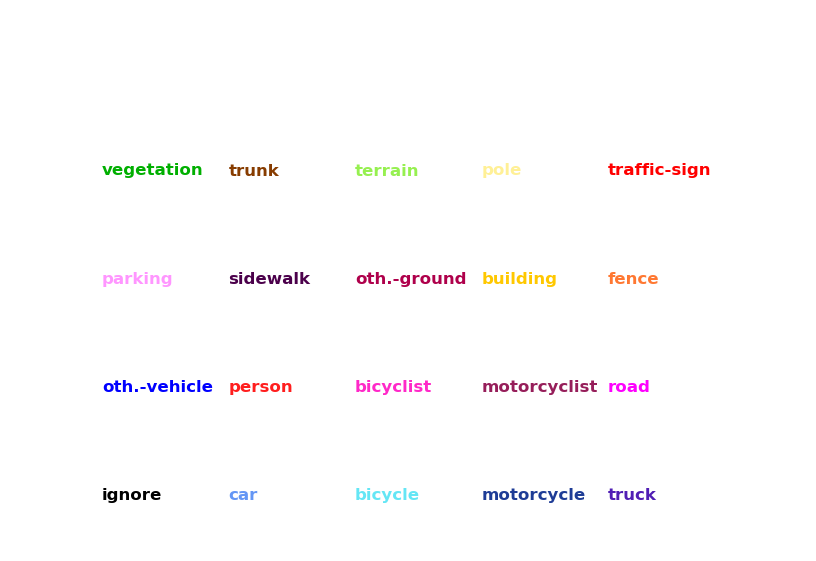}}
\vspace{5mm}
\caption{Color code used to represent each class on nuScenes (top) and SemanticKIITI (bottom) in Figure 3.}
\label{fig:color_code}
\end{figure*}

\clearpage    

\end{document}